%% file: main-v14.tex
\documentclass{article}
\usepackage{graphicx} 
\usepackage{amsmath, amsfonts, amssymb}
\usepackage{hyperref}
\usepackage{tikz}
\usepackage{booktabs}
\usepackage{xcolor}
\usepackage{float}
\usepackage[title]{appendix}
\usepackage[affil-it]{authblk}

\usepackage[table]{xcolor} 
\usepackage{array}  
\usepackage{tabularx}
\usepackage{microtype}
\usepackage{longtable}
\usepackage{authblk}
\usepackage{ragged2e}

\usepackage{pifont}
\usepackage{graphicx} 
\newcommand{\email}[1]{\thanks{Email: #1}}
\usepackage[includeheadfoot, margin=1in]{geometry}

\title{SwarmFly: A simulation platform for UAV swarm experiment design and validation}

\author[1]{Abhishek Phadke\email{abhishek.phadke@cnu.edu}}
\author[2]{Karthik Kumar Vasudeva\email{karthik.vasudeva@ttu.edu}}
\author[3]{Abhishek Joshi\email{ajoshi5@islander.tamucc.edu}}

\affil[1]{School of Engineering and Computing, Christopher Newport University}
\affil[2]{Department of Mathematics and Statistics, Texas Tech University}
\affil[3]{Department of Computer Science, Texas A\&M University-Corpus Christi}

\date{}
\begin{document}

\maketitle

\section{Abstract}

The initial development phase of UAV swarms largely depends on simulation for experimental design and validation, yet existing open-source tools are often unmaintained, have steep learning curves, or are built around a single fixed scenario. The need for a comprehensive, modular simulation platform is a recognized research gap. This paper presents SwarmFly, a MATLAB-based simulation and test platform for multi-UAV swarms that addresses these gaps. SwarmFly combines a real-time operational map, four swarm coordination modes (leader-follower, decentralized, heterogeneous relay, and heterogeneous speed), simulated IMU telemetry, and IP-based geolocation with a plugin architecture that lets researchers add behaviors, fault models, and analysis tools without touching the core code. Eight bundled plugins extend the base simulator into a full test harness. The SwarmFly platform exposes multi-agent aerial swarms to a wide range of internal and external disruptions, enabling observation and quantification of underlying swarm control and behavioral mechanisms. This study verifies and characterizes each subsystem through eight experiments that measure formation accuracy, wind tolerance, fault recovery, energy endurance, and airspace compliance. The platform runs entirely in MATLAB. Its modular design supports straightforward extension toward hardware-in-the-loop testing, larger swarms, and higher-fidelity dynamics. An open-source release is available at \href{https://github.com/abhishekphadke/SwarmFly.git}{[https://github.com/abhishekphadke/SwarmFly.git]}

\textbf{Keywords:} \textit{UAV, swarm, simulation, experiment-framework}

\section{Introduction}

The domain of UAV swarms and their applications in agriculture \cite{qu2022uav}, defense \cite{ran2025bio}, emergency communications \cite{huang2024cooperative}, search and rescue \cite{ruetten2020area}, entertainment \cite{alqudsi2026enhancing}, and other areas is increasing at a fast pace \cite{phadke2023examining}. UAV swarms are subject to a broad range of failures that can lead to crashes and damage to life and property, and subsequent mission failure \cite{xu2022failure}. It is vital to research UAV swarm resiliency features, applications, and improvements.  

However, the tools available to conduct this research have not kept pace with the problems they must address. Researchers studying swarm resiliency need to subject a coordinating group of agents to faults, environmental disturbances, energy constraints, and airspace restrictions,
and then measure how the swarm responds. In practice, this means either extending a simulator never designed for such experiments or, more commonly, building a special environment from scratch for each study. Both paths are
costly: the first fights the assumptions of the original tool, and the second sacrifices the consistency and comparability that accumulate when multiple studies share a common platform. Existing open-source swarm simulators compound the
problem. Several are no longer maintained \cite{dimmig2024survey}, others demand substantial programming investment before a first experiment can run, and most are built
around a single coordination scheme or scenario rather than a reconfigurable test harness. The result is a recognized gap: there is no lightweight, extensible, and actively maintained platform that lets a researcher inject a
fault, vary the wind, switch coordination modes, and quantify the outcome without rewriting the simulator each time.

This paper addresses that gap with SwarmFly, a MATLAB-based simulation and test platform for multi-UAV swarms. SwarmFly pairs a real-time operational core with a plugin architecture that lets researchers add behaviors, fault
models, and analysis tools without modifying the underlying code. The
contributions of this work are fourfold: (i) a modular, handle-class
simulation core that runs in real time and is deployable as a MATLAB toolbox
or in the browser; (ii) four swarm coordination modes spanning leader-follower,
decentralized, and heterogeneous configurations; (iii) a suite of eight plugins that extend the base simulator into a test harness covering fault injection,
performance metrics, energy modeling, collision avoidance, geofencing, 3D
visualization, and automated scenario regression; and (iv) a set of eight
experiments that exercise each subsystem and characterize swarm behavior
across formation, wind, fault, energy, and airspace-compliance conditions.

UAV swarm simulation platforms are integral to the experimental design and validation process. Novel algorithms, payloads, and even physical robotic platforms require extensive testing in simulation before deployment in the real world \cite{phadke2023designing}. This keeps development and hardware costs in check while reducing the risk of accidents and damage. Simulation platforms also enable rapid prototyping, automated testing, and the enforcement of a variety of conditions and faults. For example, faults such as motor freeze, GPS denial, and battery failures can cause aerial platforms to crash. It is much more viable to test these conditions in simulation multiple times before hardware tests are conducted.

As robotic platforms function in simulated space, they generate vital operational log data that can be used for various purposes. The primary function is to demonstrate proof of work for the platform being tested. Additionally, these logs can be used to create platform policies and, in some cases, to train ML models. For example, experiments conducted in vehicle movement, navigation, and communication in simulation platforms collected images and sensor outputs that were used to train models for better performance of those vehicles in the real world \cite{ju2022transferring, yu2025depth, osinski2020simulation, cai2025navdp}. Experiments on novel implementations in UAV swarms are often conducted in standalone deployments, where researchers specifically code environments, UAV agents, obstacles, communication policies, and other components, and then present results. However, it is difficult to maintain consistency in such experiments across different environments, dependency versions, and parameters. Cross-platform, regularly updated, open source simulators can help address these challenges. Platforms such as CoppeliaSim \cite{rohmer2013v} and Webots \cite{michel1998webots} \cite{michel2004cyberbotics} are good candidates to test novel implementations in robotics. However, they often have a learning curve, which can take time away from researchers. 

SwarmFly is released under the [MIT] license with current version tagged at v2.0.0 and several upgrades currently in beta. It requires MATLAB R2025b and no additional toolboxes. The eight experiments in Section~\ref{sec:experiments} ship as runnable scripts in the repository, each seeded as described so that the reported values reproduce on the same MATLAB release.

\section{Background}
\begin{figure}[H]
    \centering
    \includegraphics[width=1\linewidth]{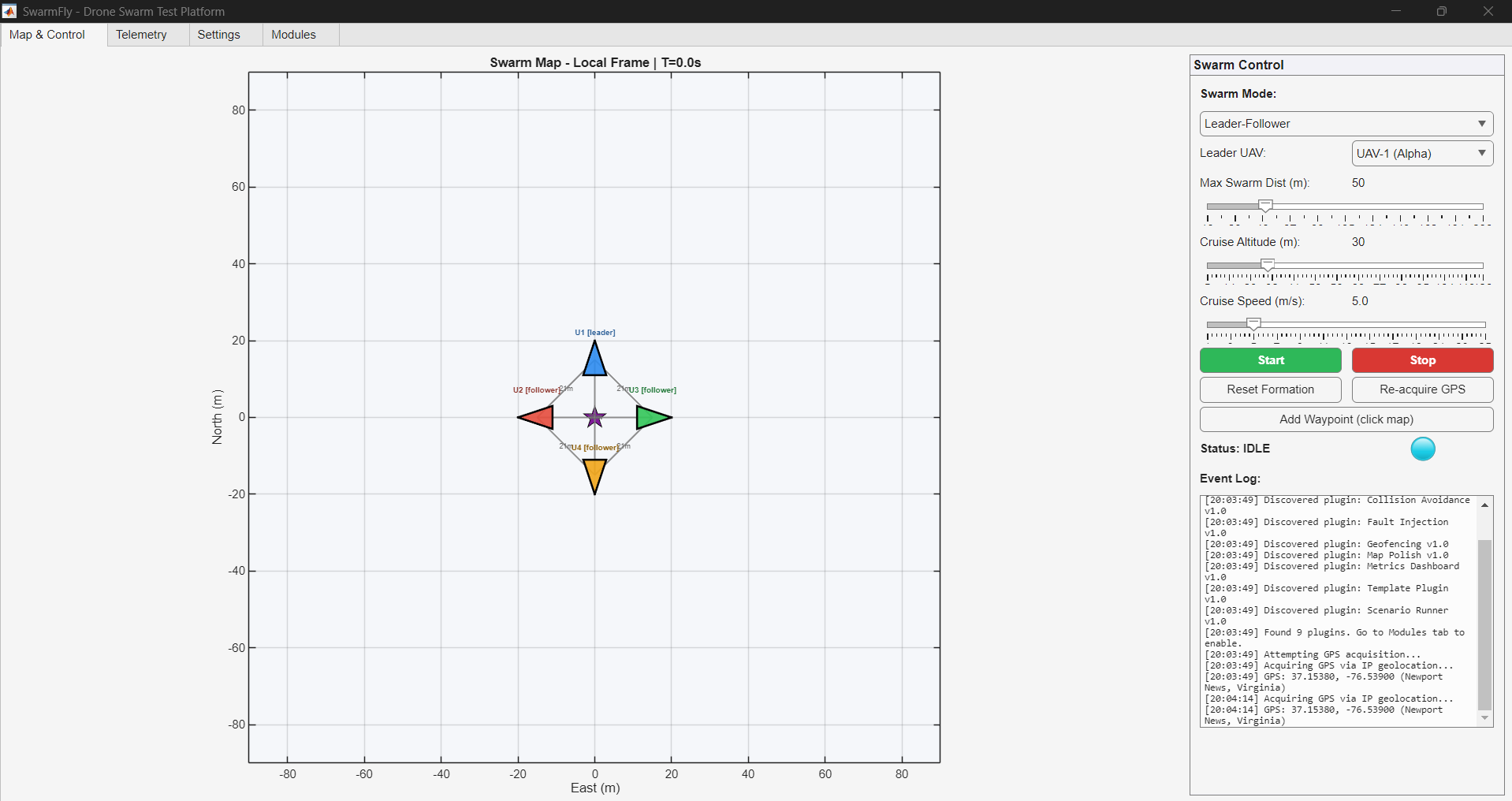}
    \caption{SwarmFly Main screen (v2.x)}
    \label{fig:main}
\end{figure}

The simulation tool presented in this study aims to address a critical gap in UAV swarm education \cite{PPR:PPR987107}, testing, and experimental validation. The study presents \textbf{SwarmFly}, a MATLAB-based simulation and test platform for multi-UAV swarms. Figure \ref{fig:main} shows the main screen for the platform.

During development, a base review and assessment of open-source swarm simulator platforms was conducted to compare and contrast the current work with existing tools \cite{robotics12020053}. Table \ref{tab:bckg} outlines existing tools along with their direct links to the code repository as of \textbf{May 31, 2026}. Although Swarmlab \cite{9340854} has not been updated in the past 6 years, it is included in the background study because several of its features have inspired SwarmFly's functionality. In addition to outlining the development of SwarmFly, this study provides researchers with additional tools, enabling them to make informed decisions about the best tools for their UAV swarm simulation requirements.  

\begin{table}[H]
\centering
\caption{List of current open source UAV Swarm simulation tools}
\label{tab:bckg}
\small
\begin{tabular}{@{}
>{\centering\arraybackslash}p{4.0cm}
>{\centering\arraybackslash}p{2.0cm}
>{\centering\arraybackslash}p{6.0cm}
>{\centering\arraybackslash}p{2.0cm}
@{}}
\toprule
\textbf{Code link \& study reference when available} &
\textbf{Tool name} &
\textbf{Simulator Purpose} &
\textbf{Program Stack} \\
\midrule
    \cite{9340854} \& \href{https://github.com/lis-epfl/swarmlab}{Link} & SwarmLab & A single and multi-drone UAV simulator with wind, obstacle, and path-planning effects & MATLAB \\
    \midrule
    \href{https://github.com/divs-spec/skysync}{Link}& skysync &  Swarm orchestration tool for turning real-time drone telemetry into coordinated multi-UAV missions & Flask, Python, MATLAB \\
    \midrule
    \cite{zhou2025uavnetsim} \& \href{https://github.com/Zihao-Felix-Zhou/UavNetSim}{Link} & UavNetSim & Model UAV communication Networks & Python \\
    \midrule
    \href{https://github.com/pabloramesc/uav-swarm-sim}{Link} & uav-swarm-sim & UAV swarm simulator to demonstrate decentralized ad-hoc networks in  autonomous multi-copters & Python, C++ \\
    \midrule
    \href{https://github.com/internashionalist/Sky_Weave}{Link} & Sky\_Weave & UAV swarm real-time networking, distributed autonomy, and telemetry visualization  & C++, Next$.$js   \\
    \midrule
    \cite{fernando2019formation} \cite{9560899} \cite{lee2010geometric} \& \href{https://github.com/malintha/multi_uav_simulator}{Link} & Mavswarm & Simulating heterogeneous quadrotor swarms with support for quadrotor control, trajectory optimization and receding horizon planning (RHP)  & C++  \\
    \midrule
    \href{https://github.com/shupx/swarm_sync_sim}{Link} & Swarm\_sync\_sim & Synchronized (lock-stepped) simulation platform for simulating various kinds of robots, including quadrotors, uncrewed ground vehicles (UGV), fixed-wing UAVs  & C++ and C  \\
    & & & \\
    & & & \\
    & & & \\

    \bottomrule
    \end{tabular}
\end{table}

\section{Simulator framework}

\label{sec:framework}

The simulator was built around two requirements: the simulation core must run fast enough for real-time visualization at 10--30\, Hz, and the system must be modular, allowing users to add new behaviors, fault models, and analysis tools without modifying the core framework. The design resolves this through a handle-class architecture with a plugin system that hooks into the simulation loop at instance points.

\subsection{Simulator architecture}
\label{sec:architecture}

SwarmFly is implemented as a single MATLAB \texttt{handle} class (\texttt{SwarmFly.m}) with all simulation state, GUI construction, and rendering logic in one file. A handle class was chosen over a value class for a practical reason: every callback, timer function, and plugin receives a reference to the same object, so writes to \texttt{app.UAVPositions} in a plugin's \texttt{onStep} function are immediately visible to the rendering code that runs a few lines later in the same tick. There is no copying, no message passing, and no synchronization overhead. The application launches a \texttt{uifigure} window whose size adapts to the host screen resolution. The GUI is organized around a \texttt{uitabgroup} with four core tabs (Map \& Control, Telemetry, Settings, and Modules), plus any additional tabs created by enabled plugins. All layouts use \texttt{uigridlayout} containers, which automatically resize when the user drags the window edges.

\subsection{Simulation loop}
\label{sec:simloop}

A \texttt{timer} object in fixed-rate mode drives the simulation. At each tick, the \texttt{simStep()} method executes the following sequence:

\begin{enumerate}
    \item Advance simulation time: $t \leftarrow t + \Delta t$, where $\Delta t = 1 / f_{\text{update}}$.
    \item Compute the wind disturbance vector from speed and direction inputs using \eqref{eq:wind}.
    \item Run the active swarm mode's physics function. This is where the core flight dynamics happen: waypoint tracking, formation control, or decentralized swarming, depending on the selected mode. Each function updates \texttt{app.UAVPositions} and \texttt{app.UAVHeadings} in place.
    \item Apply altitude convergence. A proportional controller \eqref{eq:alt_ctrl} pulls each UAV toward the cruise altitude. This runs after the mode-specific physics, so that altitude corrections do not interfere with the horizontal dynamics.
    \item Record telemetry. Position and simulated IMU readings (accelerometer, gyroscope, magnetometer) are appended to a rolling buffer of 500 samples per UAV. The IMU values are derived from position finite differences plus Gaussian noise, following the sensor models in \eqref{eq:accel_x}--\eqref{eq:noise}.
    \item Execute plugin step functions. The loop iterates over every entry in the plugin registry and calls \texttt{onStep(app)} for each enabled plugin that declared \texttt{hasStep = true}. Because plugins receive the \texttt{app} handle, they can read and write any simulation state. A \texttt{try-catch} block wraps each call so that a crashing plugin does not halt the simulation.
    \item Update map graphics. Persistent graphics handles (patches, lines, text objects) have their \texttt{XData}, \texttt{YData}, \texttt{Visible}, and \texttt{Color} properties updated to reflect the current UAV positions. No graphics objects are created or destroyed during this step.
    \item Update telemetry plots, but only when the Telemetry tab is the active tab. This conditional check avoids spending render time on six axes.
    \item Call \texttt{drawnow limitrate} to flush the graphics pipeline. The \texttt{limitrate} flag throttles actual screen redraws to roughly 20\, FPS regardless of the timer frequency, which prevents the GUI from becoming unresponsive at high update rates.
\end{enumerate}

The ordering of steps 3 through 6 is load-bearing. The physics engine writes positions first. Plugins then modify those positions (fault injection adds drift, collision avoidance pushes UAVs apart). The renderer at step~7 sees the final, post-plugin state. This means multiple plugins compose their effects naturally: a GPS drift fault and a collision avoidance correction both write to the same position array, and the map shows the combined result.

\subsection{Rendering strategy}
\label{sec:rendering}

Early versions of the simulator used a create-and-destroy pattern: every frame deleted all graphical objects via \texttt{findobj} and recreated them. This approach caused visible flickering and consumed most of the per-tick time budget. The current design creates all persistent graphics objects once during initialization (\texttt{initMapGraphics}, \texttt{initTelemetryGraphics}) and updates only their data properties each tick. For the map, this means 4 patch objects (UAV triangles), 4 text objects (labels), 4 line objects (trails), 6 line objects (pairwise connections), 6 text objects (distance labels), 1 line object (relay link), and 1 plot object (base station). In the telemetry tab, 24 line objects represent 4 UAVs across 6 subplots, with up to 3 field components each. These handles are stored in cell arrays on the app object and indexed by UAV number. The rendering cost per tick reduces to property assignment (\texttt{handle.XData = newValues}), which MATLAB handles efficiently without triggering full object reconstruction. Combined with \texttt{drawnow limitrate}, the result is smooth animation even with all plugins active.

\subsection{Swarm coordination modes}
\label{sec:modes}

SwarmFly has four coordination modes, each implemented as a separate method that \texttt{simStep} dispatches via a \texttt{switch} statement.
\begin{enumerate}
\item {Leader-Follower}

One UAV (selectable via the GUI) acts as the leader. If waypoints have been placed, the leader flies toward the next waypoint using bearing and speed calculations \cite{Doodeman_2025} from \eqref{eq:waypoint_bearing} and \eqref{eq:wp_speed}. When no waypoints are active, the leader follows a circular orbit trajectory \eqref{eq:orbit} with a 40\,m radius. The remaining three UAVs maintain a diamond formation by tracking target positions computed as the leader's position plus scaled offsets \eqref{eq:desired_pos}--\eqref{eq:offset_scale}. Follower velocity is determined by a proportional controller \eqref{eq:prop_follower} with a gain of $K_p = 2.0$ and speed limited to $1.5 \times V_c$. When the position error exceeds the maximum swarm distance, the error vector is clamped according to \eqref{eq:dist_constraint_position} to prevent followers from overshooting during recovery.

\item {Decentralized}

All four UAVs operate independently using a Reynolds-inspired Boids model \cite{machines13040255}. Each UAV's velocity is the sum of three components \eqref{eq:composite_vel}: a stochastic wander term \eqref{eq:wander} that produces smooth, individualized trajectories through per-UAV phase offsets; a separation force \eqref{eq:separation} that repels UAVs from neighbors closer than 10\,m; and a cohesion force \eqref{eq:cohesion} that pulls each UAV toward the swarm centroid \eqref{eq:centroid} with a gain of $K_c = 0.3$. Together, these three forces produce emergent flocking behavior in which the swarm remains loosely grouped without any explicit leader or formation geometry.

\item {Heterogeneous Relay}

UAVs~1--3 operate in leader-follower mode. UAV-4 is designated as a communication relay and positions itself at 40\% of the distance between the base station (origin) and the centroid of the active UAVs \eqref{eq:relay_target}. The relay's speed is capped at 3\,m/s, which keeps it relatively stationary compared to the maneuvering swarm. Wind effects on the relay are halved to simulate a hovering platform with lower drag exposure.

\item {Heterogeneous Speed}

UAV-1 operates as a fast scout at twice the cruise speed \eqref{eq:speed_diff}. UAVs~2--4 follow at 0.6 times the cruise speed in a V-formation behind the scout. The V-formation geometry is defined by angular offsets \eqref{eq:vform_angle} relative to the scout's heading, with a 25\,m separation radius \eqref{eq:vform_offset}. The speed asymmetry tests whether a slower payload group can keep formation contact with a scout that pulls ahead during long traversals. Prior work on heterogeneous swarms has shown a clear change in performance on vital metrics, such as area coverage, when agents with different speeds are part of a swarm \cite{joshi2026classification}.

\end{enumerate}

\subsection{Telemetry and sensor simulation}
\label{sec:telemetry}

\begin{figure}[H]
    \centering
    \includegraphics[width=1\linewidth]{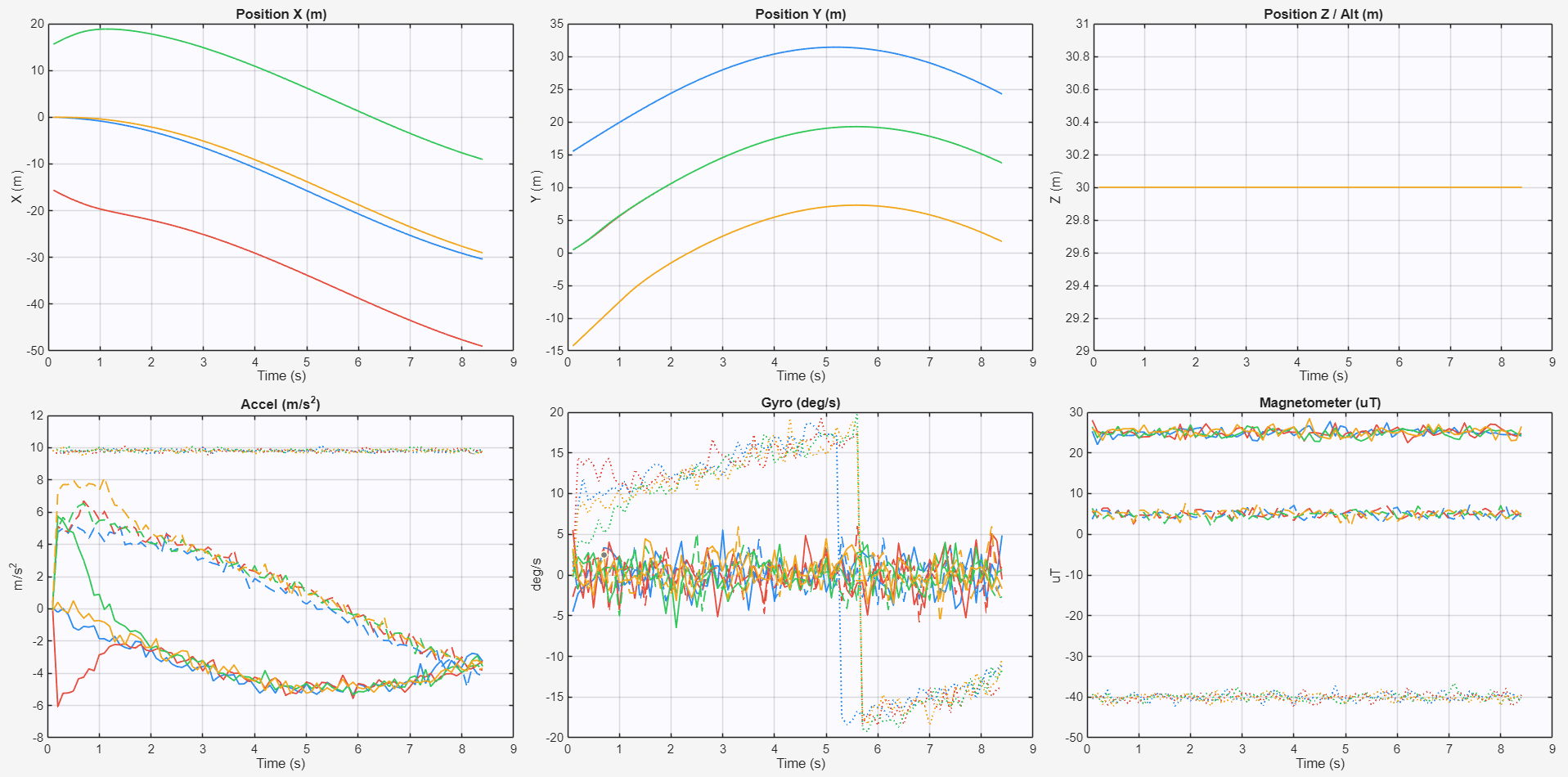}
    \caption{A view of the telemetry panel for 4 UAV agents on a standard run}
    \label{fig:tel}
\end{figure}

Each UAV generates 13 telemetry channels per tick: three position components ($x$, $y$, $z$), three accelerometer axes, three gyroscope axes, three magnetometer axes, and a timestamp. The accelerometer model \eqref{eq:accel_x} uses first-order finite differencing of position to approximate acceleration, with additive Gaussian noise ($\sigma_a = 0.3$\,m/s$^2$). The vertical channel includes a gravitational bias \eqref{eq:accel_z}. The gyroscope model produces noise-dominated output on the roll and pitch axes, with a heading-correlated signal on yaw \eqref{eq:gyro_z}. The magnetometer model \eqref{eq:mag} outputs a constant Earth-field approximation with per-axis noise. These are not high-fidelity sensor models. Their purpose is to produce telemetry data with realistic noise characteristics and plausible magnitudes, sufficient for testing data processing pipelines, plotting infrastructure, and fault detection algorithms. Researchers who need higher fidelity can replace the sensor model functions or inject real recorded data through the plugin interface. Telemetry data is stored in a struct array (\texttt{app.TelHistory}), one element per UAV, with each field as a row vector that grows each tick and is trimmed to the most recent 500 samples. The rolling buffer prevents memory from growing unboundedly during long simulation runs.
Figure \ref{fig:tel} shows an example of the standard telemetry dashboard.

\subsection{GPS integration}
\label{sec:gps}

On startup, SwarmFly attempts to geolocate the host machine by querying the \texttt{ip-api.com} REST API via MATLAB's \texttt{webread} function. If the query succeeds, the returned latitude and longitude are stored as the base station origin and displayed in the map title. The simulation operates in a local East-North-Up (ENU) coordinate frame centered at this origin, with all positions expressed in meters. If the API call fails (no internet connection, a blocked endpoint), the application defaults to coordinates near Newport News, Virginia (37.0\textdegree N, 76.0\textdegree W) and operates in local-frame mode. GPS can be reacquired at any time via a button on the control panel.

\subsection{Plugin architecture}
\label{sec:plugins}

The plugin system is the primary mechanism for extending SwarmFly without modifying the core application. Plugins are ordinary MATLAB function files that follow the naming convention (\texttt{swf\_*.m}) and return a struct that describes their capabilities and provides callback function handles.

\begin{enumerate}
\item {Plugin Discovery}

At startup, the \texttt{discoverPlugins()} method scans the \texttt{plugins/} subfolder for files matching \texttt{swf\_*.m}. Each file is called as a function, and the returned struct is validated for required fields (\texttt{id}, \texttt{name}) and augmented with defaults for optional fields (\texttt{description}, \texttt{version}, \texttt{hasTab}, \texttt{hasStep}, callback handles). The struct is appended to a registry array, and the corresponding \texttt{PluginEnabled.(id)} flag is set to \texttt{false}. If a file throws an error when called, the error is logged, and the file is skipped.

\item {Plugin Lifecycle}

Enabling a plugin triggers three actions in sequence. First, \texttt{onLoad(app)} is called, where the plugin typically initializes its state using \texttt{app.setState(id, key, value)}. Second, if the plugin declared \texttt{hasTab = true}, a new \texttt{uitab} is created in the tab group and passed to the plugin's \texttt{buildTab(app, tab)} callback, where the plugin constructs its own GUI with whatever controls and plots it needs. Third, the plugin's \texttt{onStep} handle is now eligible for execution during the simulation loop. Disabling a plugin reverses these steps: \texttt{onUnload(app)} is called for cleanup, the tab is deleted, and the state namespace is cleared.

\item {State Management}

Each plugin gets an isolated namespace within \texttt{app.PluginStates}, a nested struct keyed by plugin ID. The API has two methods: \texttt{setState(id, key, value)} and \texttt{getState(id, key)}. This keeps plugin data separated from both the core simulation state and other plugins. The namespace is automatically created on enable and destroyed on disable. One constraint that bit us early: plugin IDs must be valid MATLAB struct field names, which means no leading digits and no spaces. A plugin with the ID \texttt{3d\_view} caused MATLAB to reject the field name entirely. Renaming it to \texttt{view3d} fixed the problem.

\item {Plugin-to-GUI Communication}

Plugins that need to update their GUI elements from within \texttt{onStep} face a practical problem: the tab handle and its children are created in \texttt{buildTab}, but \texttt{onStep} runs in the timer callback context. The solution is tagging UI elements with unique strings during construction (\texttt{'Tag', 'my\_unique\_tag'}) and locating them at runtime with \texttt{findobj(app.Fig, 'Tag', 'my\_unique\_tag')}. This is wrapped in \texttt{try-catch} because the figure may be closing when the timer fires.

\end{enumerate}


\section{Simulator features and use cases}
\label{sec:features}

The base simulator provides the features outlined in the section above. The plugin capability allows users to load conditional modules for additional testing and framework validation. SwarmFly ships with eight plugins that collectively turn the base simulation into a test platform. This section describes each plugin's purpose, implementation approach, and the scenarios in which it is most useful. All plugins follow the architecture described in Section~\ref{sec:plugins} and can be turned on or off independently at runtime.

\subsection{Fault injection}
\label{sec:feat_faults}

The fault injection plugin lets researchers intentionally inject faults during a running simulation. The user selects a fault type, picks a target UAV (or all four), sets a duration and intensity, and clicks inject. The plugin stores an active-faults list in its state namespace and processes it every tick in \texttt{onStep}.

\begin{figure}[H]
    \centering
    \includegraphics[width=1\linewidth]{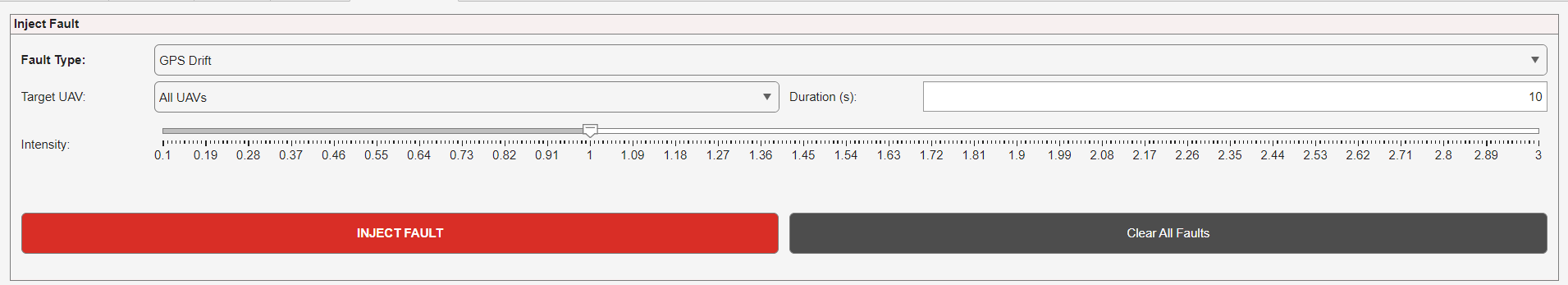}
    \caption{Fault injection plugin settings}
    \label{fig:fault}
\end{figure}

Figure \ref{fig:fault} shows the settings screen for the plugin.

The plugin implements eight fault types, each targeting a different aspect of the UAV state:

\begin{itemize}
    \item \textbf{GPS drift} adds a sinusoidal position bias that grows with elapsed time \eqref{eq:gps_drift}. The drift starts small and becomes increasingly obvious, which helps test whether a formation controller notices gradual sensor degradation.
    \item \textbf{GPS denied} replaces position updates with a random walk \eqref{eq:gps_denied}, simulating a UAV that has lost satellite lock and is flying on dead reckoning with accumulating error.
    \item \textbf{Motor failure} forces altitude decay \eqref{eq:motor_fail} at a rate proportional to the intensity parameter. The UAV loses altitude while maintaining horizontal motion, simulating partial thrust loss.

    \item \textbf{Comm blackout} injects large random position perturbations, simulating a UAV that can no longer receive formation correction commands and drifts unpredictably.
    \item \textbf{Sensor noise spike} corrupts the accelerometer telemetry with extreme noise \eqref{eq:sensor_spike}, useful for testing how IMU-dependent algorithms handle outlier readings.
    \item \textbf{Frozen actuator} locks the heading and makes the UAV drift in a straight line \eqref{eq:frozen}, regardless of controller commands.
    \item \textbf{Wind gust} applies a sudden random-direction impulse \eqref{eq:gust}, simulating turbulence or wake effects \cite{aerospace11030237}.
    
    \item \textbf{Battery critical} forces a controlled descent \eqref{eq:motor_fail} at a higher rate than motor failure, simulating an emergency landing triggered by low charge.
\end{itemize}

Each fault has an adjustable intensity multiplier ($I \in [0.1, 3.0]$) that scales the severity. Faults expire automatically after their specified duration, and the plugin's GUI displays a live table showing each active fault's type, target, remaining time, and intensity. Multiple faults can be active simultaneously across different UAVs, enabling compound failure testing. A typical experiment looks like this: inject a GPS denial on UAV-3 during a leader-follower waypoint mission and observe whether the remaining UAVs compensate, whether the formation degrades gracefully, and how long the swarm takes to recover after the fault clears.

\subsection{Performance metrics}
\label{sec:feat_metrics}

\begin{figure}[H]
    \centering
    \includegraphics[width=1\linewidth]{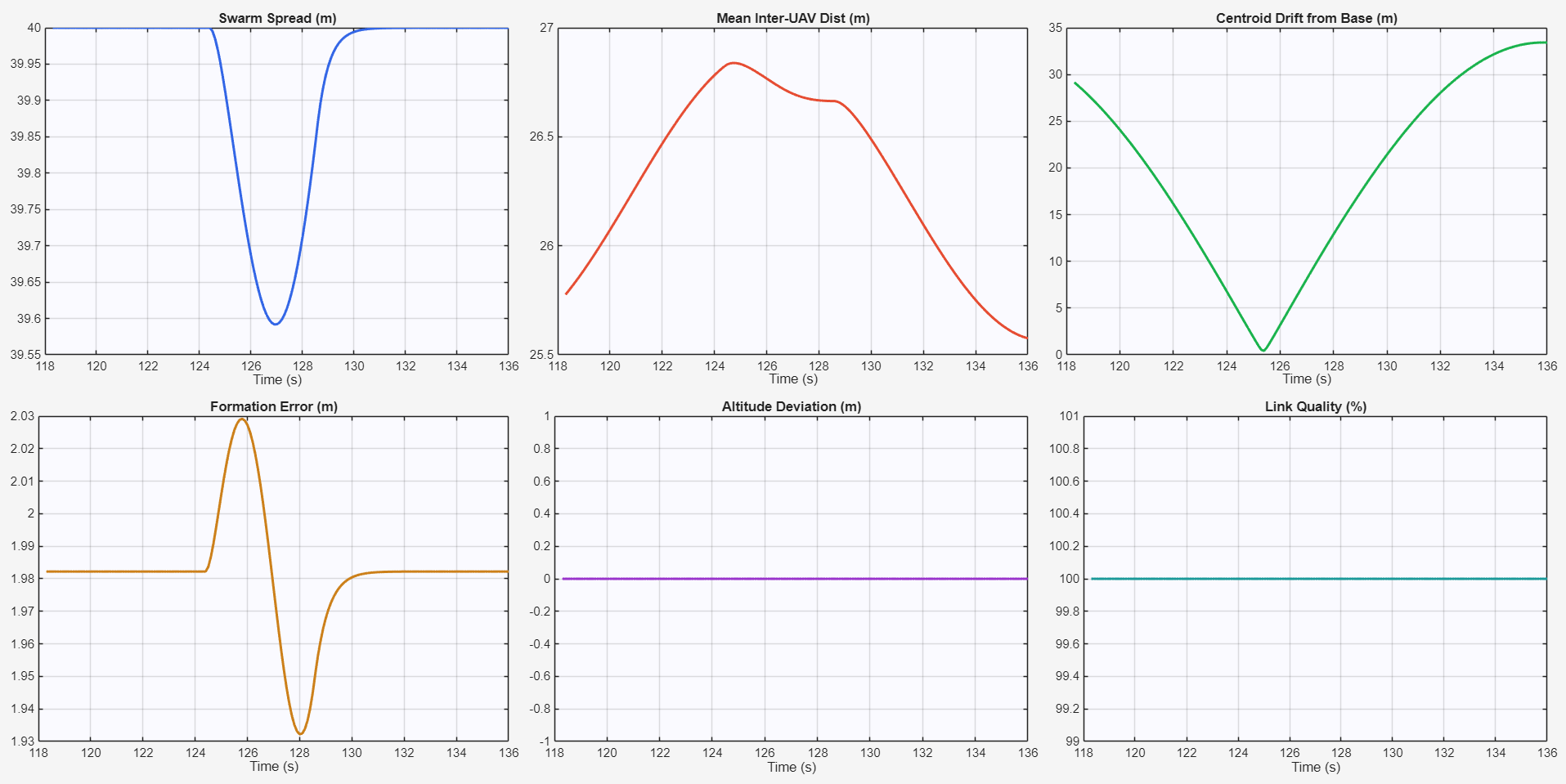}
    \caption{Performance metrics dashboard}
    \label{fig:metrics}
\end{figure}

\begin{figure}
    \centering
    \includegraphics[width=1\linewidth]{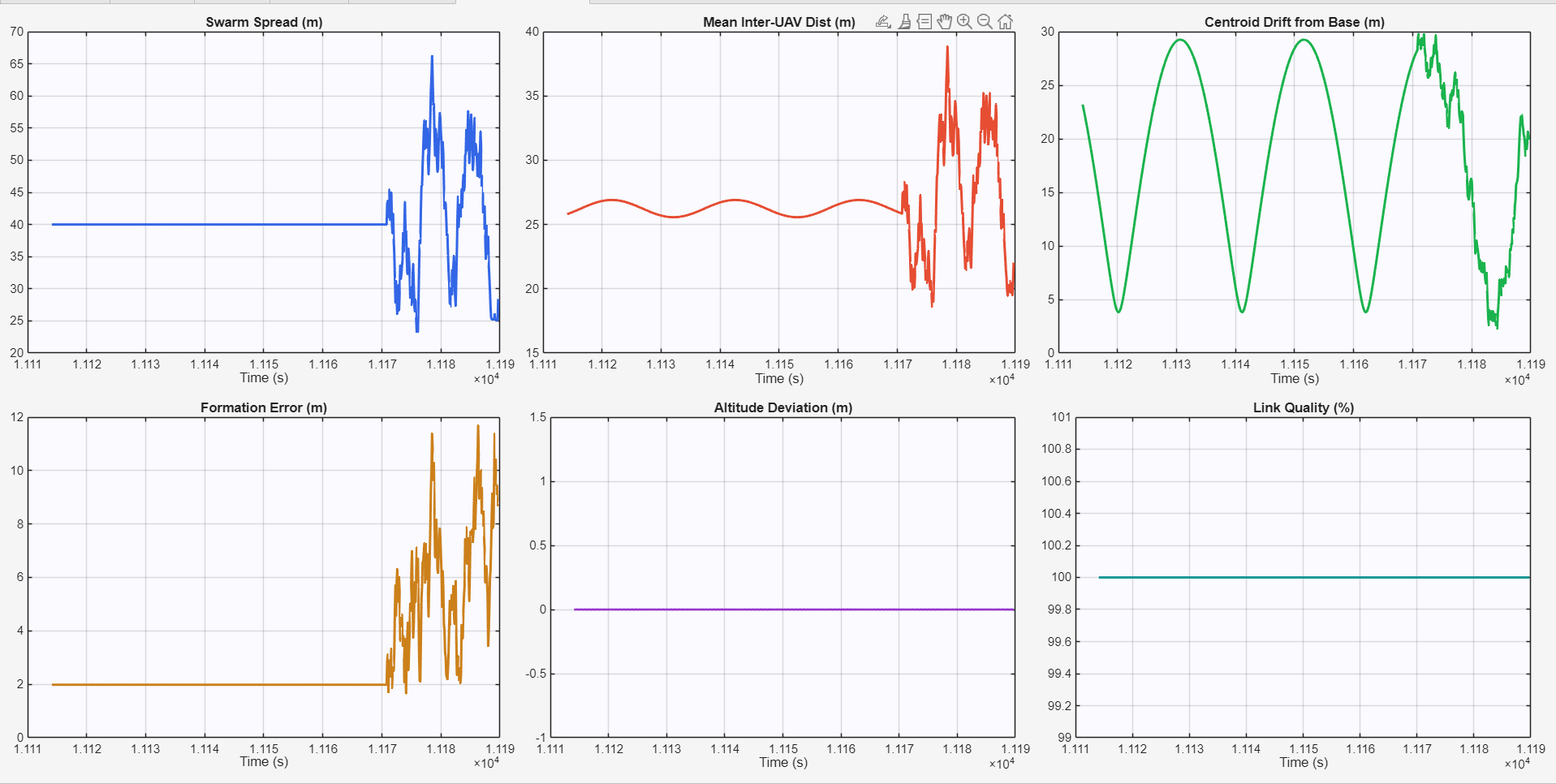}
    \caption{Changes in swarm status post fault injection}
    \label{fig:postFaultPerformance}
\end{figure}

The metrics plugin computes six KPIs every tick and plots them on dedicated axes in its tab. These are the numbers you would report in a paper to characterize swarm behavior quantitatively. \textbf{Swarm spread} \eqref{eq:spread} is the maximum pairwise distance between any two UAVs. It captures how far the swarm stretches in its worst case. \textbf{Mean inter-UAV distance} \eqref{eq:mean_dist} averages all six pairwise distances, giving a smoother measure of overall dispersion. \textbf{Centroid drift} \eqref{eq:centroid_drift} tracks how far the swarm's center of mass has moved from the base station, which matters for relay coverage analysis. \textbf{Formation error} \eqref{eq:form_error} measures how well followers maintain their assigned offsets from the leader, expressed as the mean positional deviation from ideal positions. \textbf{Altitude deviation} \eqref{eq:alt_dev} captures the average absolute difference between actual and commanded altitude across all UAVs. \textbf{Link quality} \eqref{eq:link_quality} reports the percentage of UAV pairs that are within communication range. These metrics are buffered to 1,000 samples and plotted as rolling time series. They update only when the metrics tab is visible, using the same conditional rendering approach as the core telemetry plots. Combined with the fault injection plugin, this creates a quantitative test workflow: inject a fault, observe the metric response in real time, and use the time-series data to compute recovery times, maximum excursions, and steady-state error values. Figure \ref{fig:metrics} shows a view of the metrics dashboard when the plugin is enabled. Figure \ref{fig:postFaultPerformance} shows changes in swarm status and performance after a fault is injected.

\subsection{Battery and energy model}
\label{sec:feat_battery}

\begin{figure}
    \centering
    \includegraphics[width=1\linewidth]{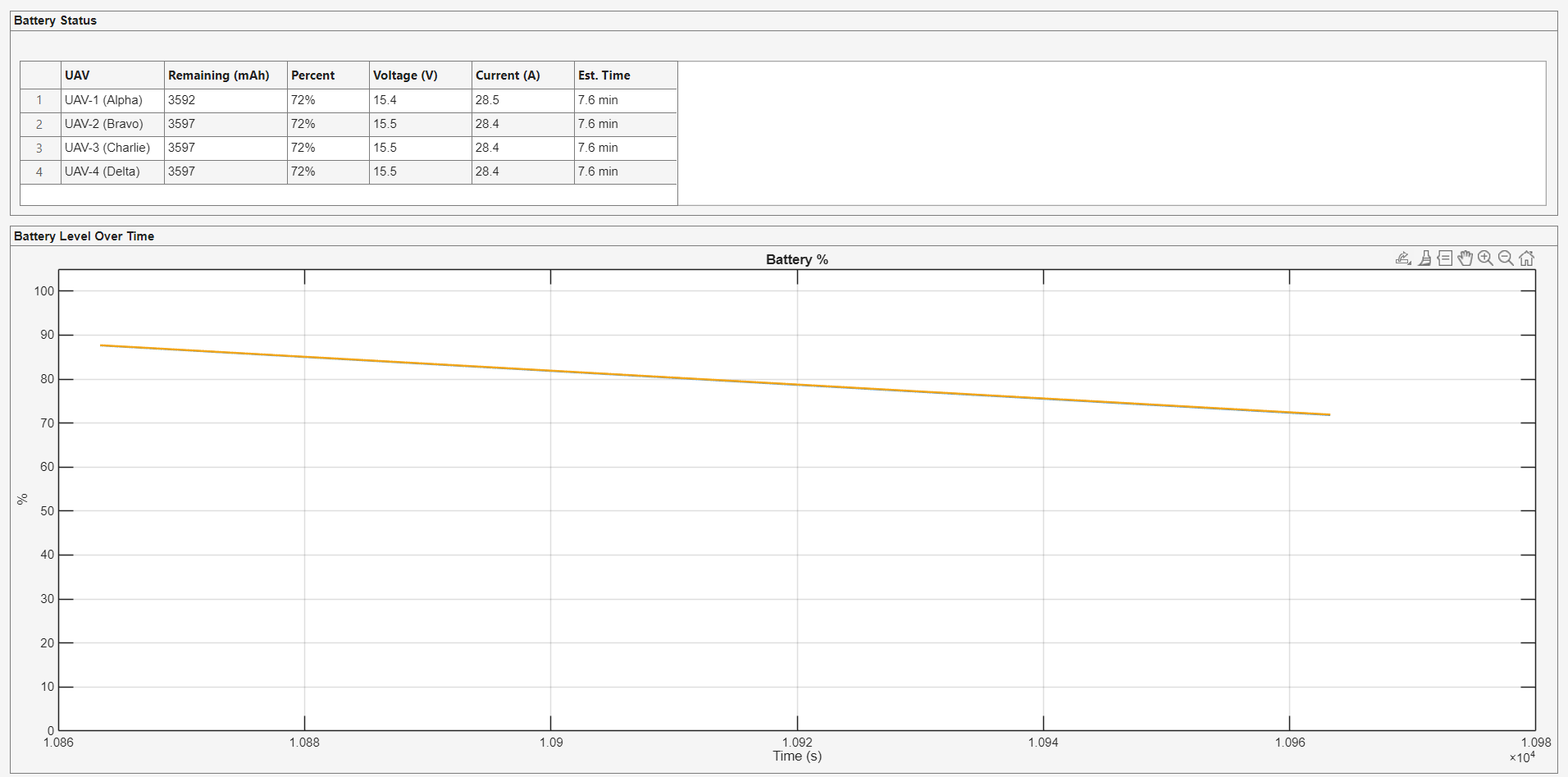}
    \caption{Battery status plugin GUI}
    \label{fig:batt}
\end{figure}

Real mission planning cannot ignore energy. The battery plugin simulates a 4S LiPo battery for each UAV using a first-order current-draw model \eqref{eq:current} that accounts for hover load ($I_{\text{hover}} = 18$\,A), forward-flight load ($K_v = 2.0$\,A$\cdot$s/m), altitude penalty ($K_h = 0.05$\,A/m above 20\,m), and wind resistance ($K_w = 0.5$\,A$\cdot$s/m). Total capacity is 5,000\, mAh per UAV. Battery drain follows \eqref{eq:battery_drain}, and voltage sag is modeled as a linear function of remaining capacity \eqref{eq:voltage}, ranging from 16.8\, V at full charge to 12.0\, V at depletion. The plugin's GUI shows a per-UAV status table with remaining mAh, percentage, voltage, current draw, and estimated remaining flight time \eqref{eq:endurance}. A historical plot tracks battery percentage over time for all four UAVs. Warnings are logged at 20\% and 10\% remaining capacity. The model does not currently force a landing at depletion, but the Battery Critical fault type in the fault injection plugin can simulate that behavior. In practice, the energy model answers questions like: can a 10-minute waypoint mission with 8\,m/s winds complete with 20\% reserve across all UAVs? This provides an early-stage endurance feasibility check. The battery plugin GUI is shown in Figure~\ref{fig:batt}.

\subsection{Collision avoidance}
\label{sec:feat_collision}

The collision avoidance plugin adds a safety layer that runs after the physics engine and before rendering. Every tick, it computes pairwise distances for all UAV combinations and applies a repulsive force \eqref{eq:repulsion} when any pair falls below the safe distance (default 5\,m). The force is proportional to the penetration depth and acts along the separation vector \eqref{eq:sep_vec}, pushing both UAVs apart symmetrically \eqref{eq:col_correction}. The plugin tracks two counters: near-misses (distance below safe threshold) and collisions (distance below 40\% of the safe threshold). Its GUI shows these counts, a slider to adjust the safe distance from 1--20\,m, and a rolling plot of the minimum pairwise distance over time with a threshold line. The collision system gets a harder workout when combined with fault injection. Injecting a frozen actuator fault into one UAV in a tight formation and observing whether the separation forces prevent impact provides a concrete, measurable safety test.

\subsection{Automated test scenarios}
\label{sec:feat_scenarios}

\begin{figure}[H]
    \centering
    \includegraphics[width=1\linewidth]{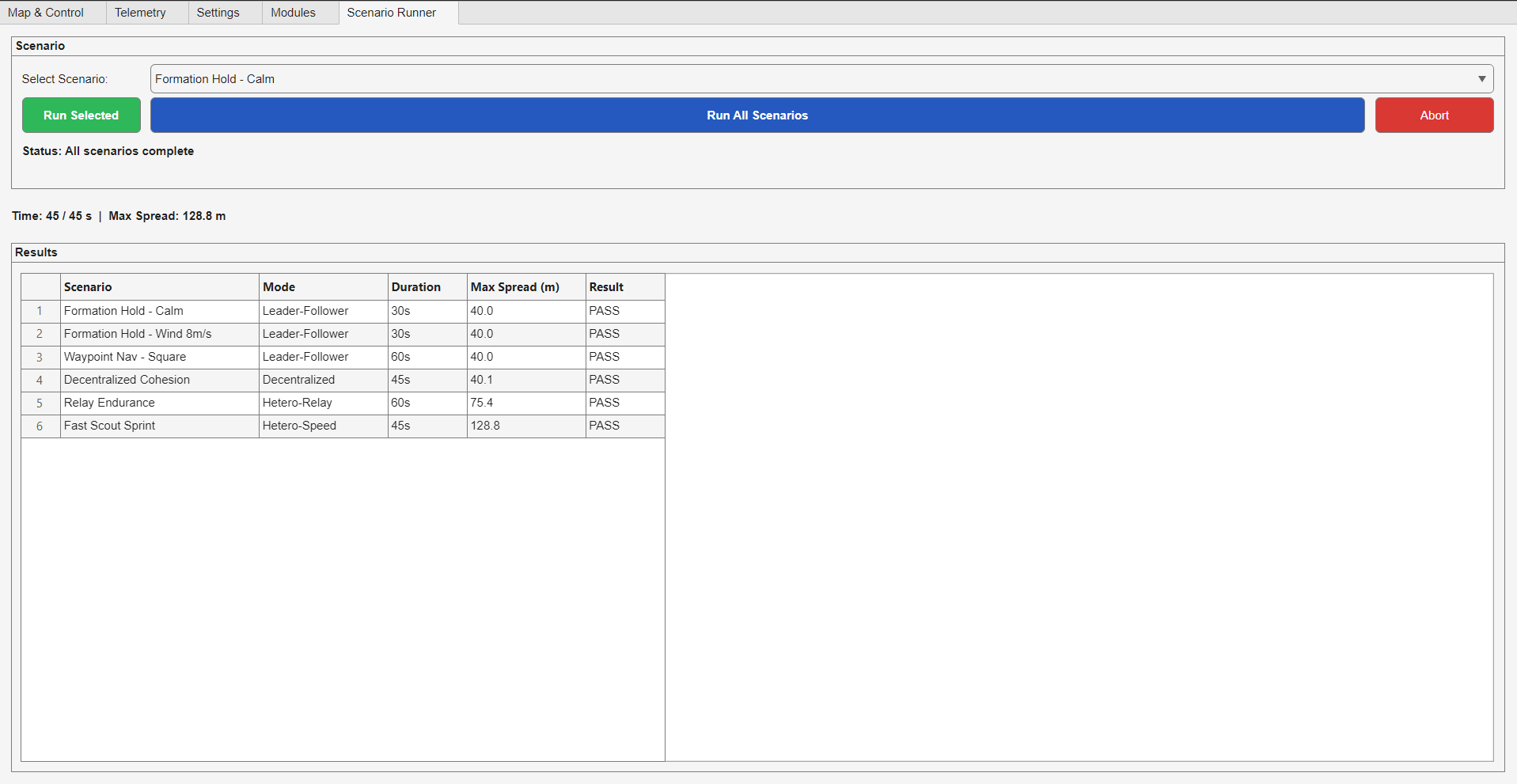}
    \caption{Automated scenario testing plugin}
    \label{fig:autoTest}
\end{figure}

The scenario runner plugin defines six test profiles as struct arrays and executes them sequentially with automated setup, timing, and pass/fail evaluation. Each scenario specifies a swarm mode, waypoint sequence, wind conditions, duration, and maximum allowable swarm distance. When a scenario starts, the plugin calls the core app's \texttt{onReset}, configures the mode and parameters, loads waypoints, and starts the simulation. During execution, it tracks the maximum observed swarm spread. When the timer expires, it stops the simulation and evaluates whether the spread stayed below twice the configured max distance. The six built-in scenarios cover the main operating conditions: calm and windy formation hold, waypoint navigation through a square pattern, decentralized cohesion under crosswind, relay endurance during an extended traverse, and a fast-scout sprint testing speed-differential coordination. Running all six in sequence with the ``Run All'' button produces a results table with pass/fail verdicts, which gives a repeatable regression-style test suite. Adding new scenarios requires only appending to the struct array in the plugin's \texttt{onLoad} function. No GUI changes are needed because the dropdown populates itself from the scenario names. Figure \ref{fig:autoTest} shows the automated scenario tester plugin with 6 scenarios tested and their results.

\subsection{3D Visualization}
\label{sec:feat_3d}

\begin{figure}[H]
    \centering
    \includegraphics[width=1\linewidth]{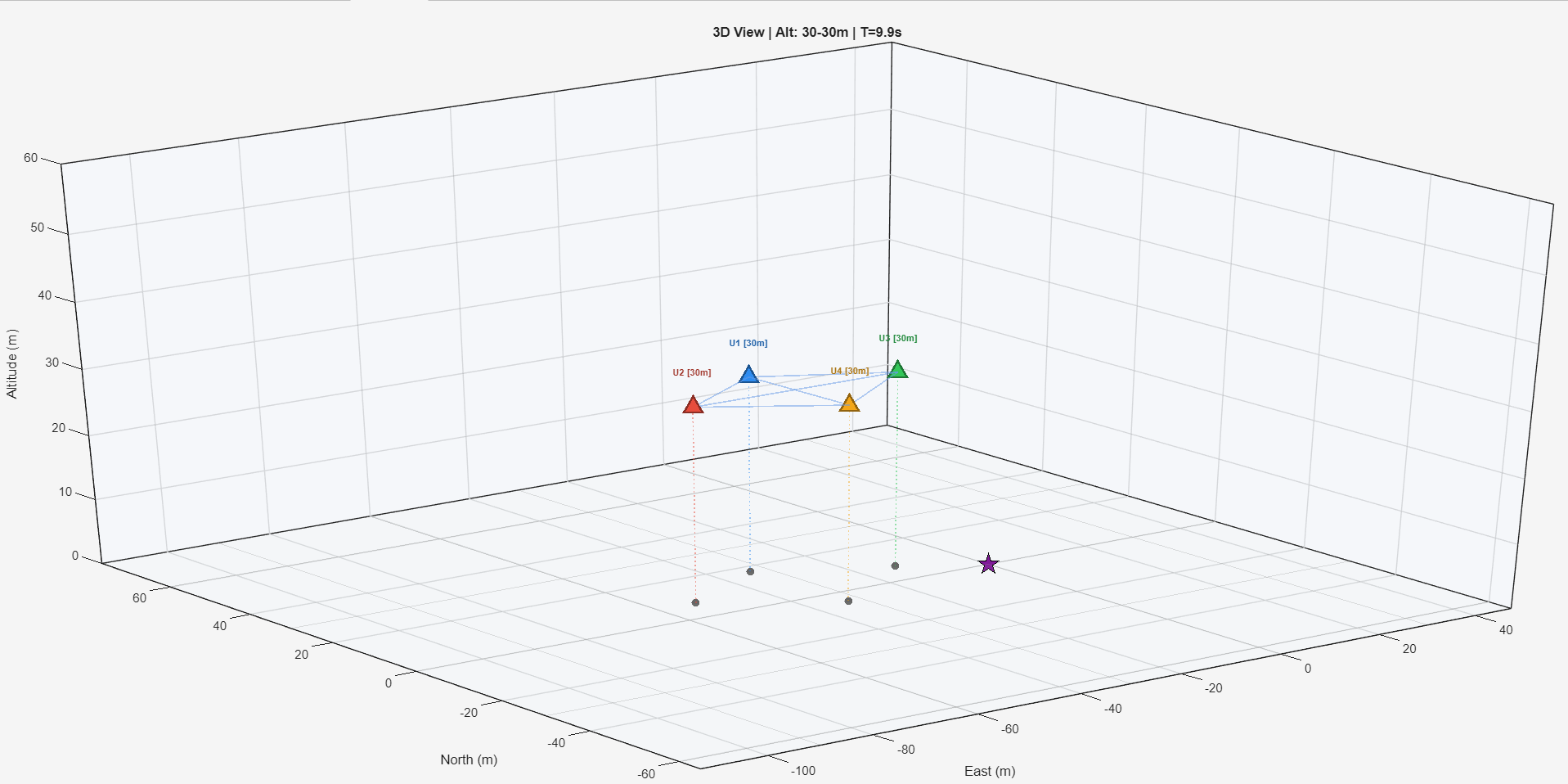}
    \caption{3D viewer plugin}
    \label{fig:3d}
\end{figure}

The 2D map is efficient for monitoring position and connectivity, but it hides the altitude dimension. The 3D view plugin creates a \texttt{uiaxes} with perspective projection and draws each UAV as a colored triangle marker at its full $(x, y, z)$ position. Dotted altitude stems drop from each marker to a transparent ground-plane mesh at $z = 0$, giving an immediate sense of how high each UAV is flying. Gray shadow markers on the ground indicate the plan view projection. Connection lines between UAVs are drawn in 3D and color-coded: blue for healthy links, red for pairs that exceed the maximum distance. Each label shows the UAV number and current altitude. The 3D axes support MATLAB's built-in rotate and zoom interactions, allowing the user to view the swarm from multiple angles. The plugin updates only when its tab is visible, since 3D rendering is more intensive than 2D property updates. Figure \ref{fig:3d} shows a view of the 3d visualization plugin screen. 

\subsection{Geofencing and no-fly zones}
\label{sec:feat_geofence}

\begin{figure}[H]
    \centering
    \includegraphics[width=1\linewidth]{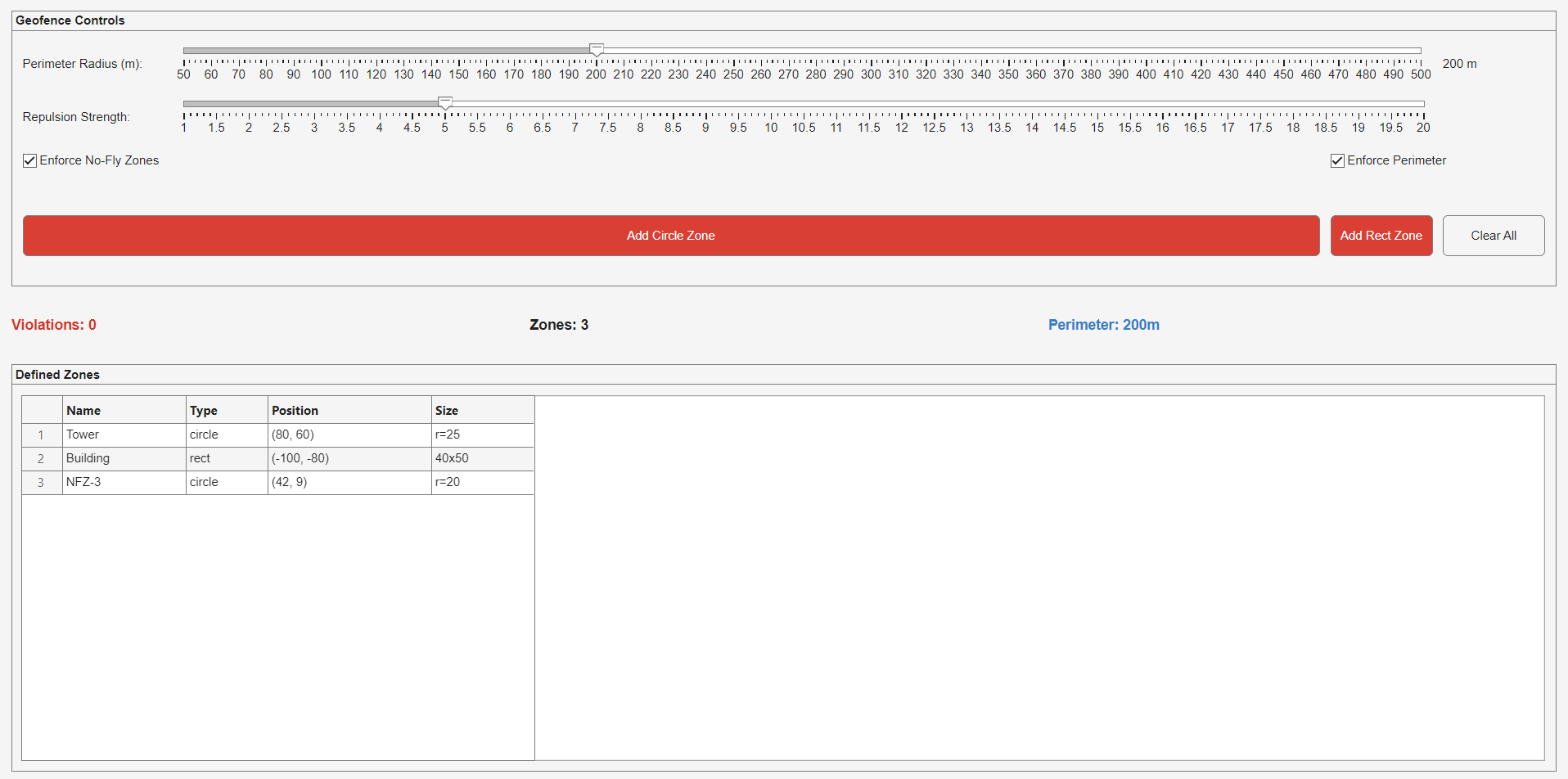}
    \caption{Geofence plugin view}
    \label{fig:geofence}
\end{figure}

\begin{figure}[H]
    \centering
    \includegraphics[width=1\linewidth]{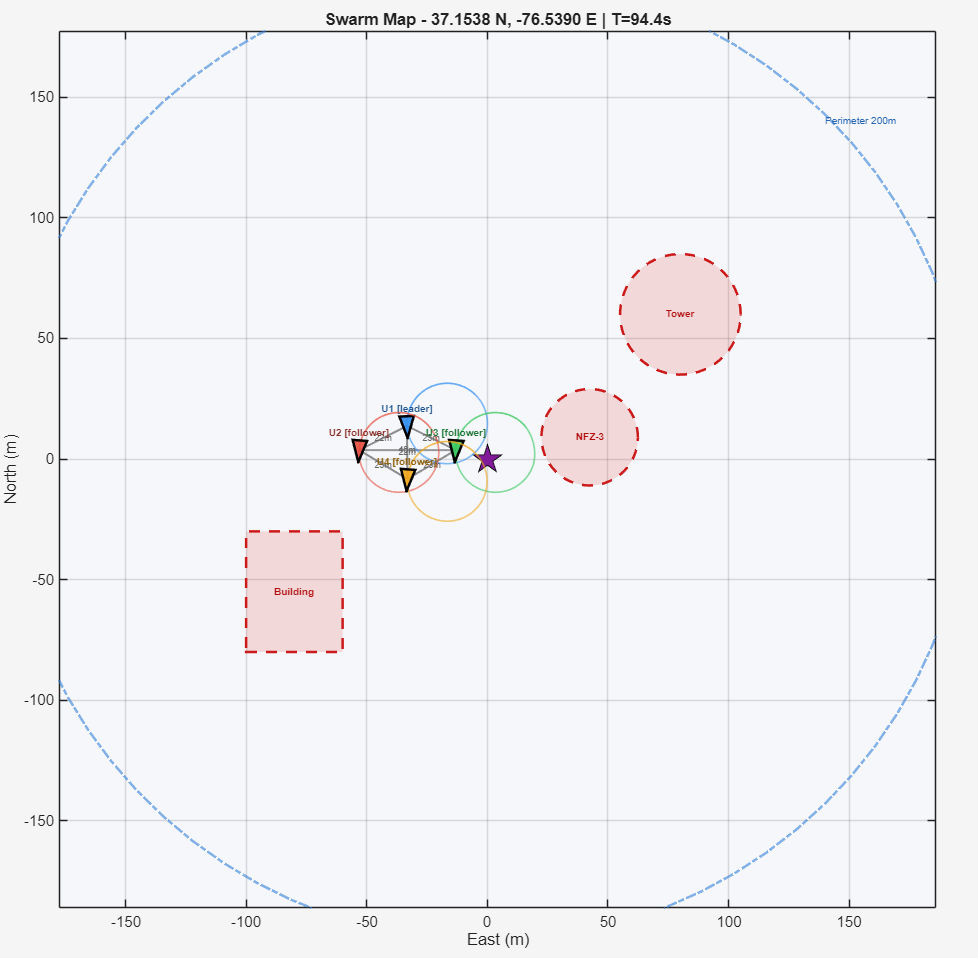}
    \caption{No-fly zones and perimeter fence rendered on the main map.}
    \label{fig:effect}
\end{figure}

The geofencing plugin adds spatial constraints to the simulation. It supports two zone types: circular (defined by a center and a radius) and rectangular (defined by a corner and dimensions). Zones appear on the main map as red dashed shaded regions. The plugin ships with two demo zones, and new zones can be placed interactively by clicking on the map. A perimeter fence defines the maximum operating radius from the base station, displayed as a dashed blue circle on the map. Figure \ref{fig:geofence} shows the Geofence plugin console with 2 default zones and one custom zone added. Figure \ref{fig:effect} shows the zones rendered on the main map.

The default radius is 200\,m. Enforcement is physics-based. When a UAV enters a no-fly zone, a repulsive force \eqref{eq:zone_force} pushes it outward along the radial direction from the zone center. A soft boundary region \eqref{eq:soft_boundary} extends 8\,m beyond each zone edge, applying a gentler force that steers UAVs away before they actually penetrate the zone. The perimeter fence applies a similar inward force \eqref{eq:perim_force} when UAVs exceed the maximum radius, with a gentle pre-warning zone at 90\% of the radius. The plugin's tab provides sliders for the perimeter radius and repulsion strength, enforcement toggles for zones and the perimeter independently, buttons to add and clear zones, a violation counter, and a table listing all defined zones with their geometry. For restricted-airspace scenarios, the test is straightforward: place a circular no-fly zone around a tower, run the swarm through a waypoint sequence that passes nearby, and check whether the formation deforms to avoid the zone and how quickly it recovers its shape afterward.

\subsection{Map visual polish}
\label{sec:feat_polish}

\begin{figure}[H]
    \centering
    \includegraphics[width=0.75\linewidth]{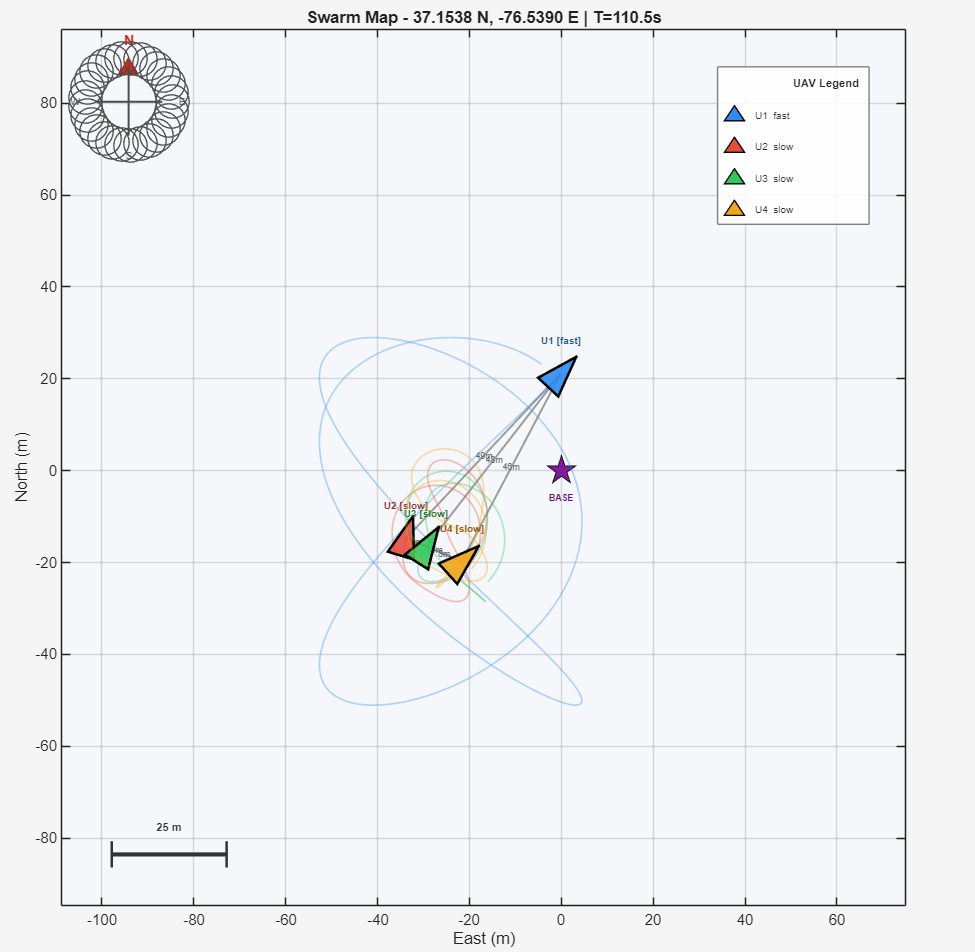}
    \caption{Map polish overlay showing compass, scale bar, legend, and base-station marker.}
    \label{fig:polish}
\end{figure}

The map polish plugin adds four overlay elements to the main map that reposition themselves each tick to stay anchored in the view corners as the axes pan and zoom: A compass rose in the top-left corner shows a red north indicator with N/S/E/W labels. A scale bar in the bottom-left corner displays a line with endcaps and a round-number distance label that auto-selects from the set $\{5, 10, 20, 25, 50, 100, 200, 500\}$\,m using \eqref{eq:scale_bar}. A semi-transparent legend box in the top-right corner shows each UAV's color, number, and current role. A ``BASE'' label marks the base station origin. Figure \ref{fig:polish} shows these additions on the simulation map.

\subsection{Extensibility}
\label{sec:extensibility}

The plugin interface is intentionally minimal. A new plugin is a single function file that returns a struct with two required fields (\texttt{id}, \texttt{name}) and up to four optional callbacks (\texttt{onLoad}, \texttt{buildTab}, \texttt{onStep}, \texttt{onUnload}). A template file (\texttt{swf\_plugin\_template.m}) ships with the platform and includes commented documentation of every available app property and method. Adding a new swarm coordination mode requires changes to the core file but is localized to four points: the dropdown item list, the mode callback, a new step function, and the dispatch statement in \texttt{simStep}. Adding more UAVs requires expanding the position, heading, color, and name arrays and adjusting the graphics initialization loops. The platform does not currently support hardware-in-the-loop (HIL) communication or ROS/MAVLink bridges, but the plugin architecture provides a clear attachment point. A plugin with \texttt{hasStep = true} could read from a serial port or TCP socket in its \texttt{onStep} callback and write received positions to \texttt{app.UAVPositions}, 
effectively replacing the simulated dynamics with live telemetry while keeping the visualization, metrics, and fault injection layers intact. This is a described attachment point, not an implemented or tested feature in the current release.

Table~\ref{tab:use_cases} summarizes how the plugins combine for common research and testing workflows.

\begin{table}[H]
\centering
\caption{Common use cases and the plugin combinations that support them.}
\label{tab:use_cases}
\small
\begin{tabular}{@{}p{4.5cm}p{4cm}p{5.5cm}@{}}
\toprule
\multicolumn{1}{c}{\textbf{Use Case}} &
\multicolumn{1}{c}{\textbf{Plugins Required}} &
\multicolumn{1}{c}{\textbf{Workflow}} \\
\midrule
\midrule
Formation resilience testing & Fault Injection + Metrics & Inject faults during flight; measure formation error and recovery time \\
Mission endurance analysis & Battery + Metrics & Run a full waypoint mission; check remaining energy at completion \\
Safety verification & Collision Avoidance + Fault Injection & Inject actuator faults in tight formations; count near-misses \\
Airspace compliance & Geofencing + Metrics & Define restricted zones; measure violations and formation deformation \\
Regression testing & Scenario Runner + Metrics & Run all scenarios after code changes; compare pass/fail results \\
Algorithm benchmarking & Custom plugin + Metrics & Replace step functions via plugin; compare KPIs against baseline modes \\
Publication figures & Map Polish + 3D View & Enable visual overlays; capture screenshots with compass, legend, scale bar \\
\bottomrule
\end{tabular}
\end{table}

\section{Experiments and results}
\label{sec:experiments}

Several experiments were conducted on the SwarmFly platform to verify each major subsystem and characterize swarm behavior across operating conditions. Each experiment checks that a subsystem behaves consistently with its governing equations; this is verification and behavioral characterization rather than validation against external flight data, which we identify as future work.

All experiments used the default 4-UAV configuration at a 10\,Hz update rate. Unless stated otherwise, cruise altitude was 30\,m, cruise speed was 5\,m/s, maximum swarm distance was 50\,m, and communication range was 200\,m. Each experiment was repeated for $n = 10$ independent trials, with trial $t$ of experiment $E$ seeded by \texttt{rng}$(1000E + t)$, so every reported value is reproducible. All runs used MATLAB R2025b on Windows 11; fixed seeds make each run reproducible on the same MATLAB release and operating system.

Two properties of the platform shape how the results are reported. First, the coordination modes and the proportional controllers are deterministic. The simulated IMU output \eqref{eq:accel_x}--\eqref{eq:noise} carries Gaussian noise, but those channels are telemetry outputs that do not feed back into the position dynamics, and every reported metric is computed from positions. Second, randomness reaches the metrics only through fault models that inject a random process: GPS denial \eqref{eq:gps_denied}, comm blackout, and wind gusts \eqref{eq:gust}. 

Deterministic experiments were executed with the same seeding protocol, but because the reported metrics are identical to numerical precision, Tables~\ref{tab:exp1}--\ref{tab:exp3} and~\ref{tab:exp6}--\ref{tab:exp8} report one representative run. Experiments~4 and~5 carry a stochastic fault component and are reported as mean $\pm$ standard deviation over the 10 trials (Tables~\ref{tab:exp4} and~\ref{tab:exp5}).

\begin{longtable}[H]{@{}c >{\RaggedRight}p{2.8cm} >{\RaggedRight}p{3.2cm} >{\RaggedRight}p{6.5cm} c@{}}
\caption{Experiment scenarios with objective and setup descriptions}
\label{tab:exp_overview} \\
\toprule
\textbf{\#} & \textbf{Experiment} & \textbf{Objective} & \textbf{Setup} & \textbf{Results} \\
\midrule
\endfirsthead
\toprule
\textbf{\#} & \textbf{Experiment} & \textbf{Objective} & \textbf{Setup} & \textbf{Results} \\
\midrule
\endhead
\midrule
\multicolumn{5}{r}{\textit{Continued on next page}} \\
\endfoot
\bottomrule
\endlastfoot

1 & Formation Hold Baseline \hspace{2cm} & Establish baseline formation performance in Leader-Follower mode under calm conditions (zero wind). & The leader followed a circular orbit \eqref{eq:orbit} at 40\,m radius with no waypoints. All four UAVs started in the default diamond formation. The simulation ran for 120\,s. & Table~\ref{tab:exp1} \\
\addlinespace

2 & Wind Resilience & Measure formation behavior under increasing wind speed. & leader-follower mode with the same orbit pattern as Experiment~1. Wind direction was fixed at 45\textdegree. Wind speed was varied from 0 to 15\,m/s in 3\,m/s increments. Each condition ran for 60\,s; metrics were recorded from $t = 10$\,s. & Table~\ref{tab:exp2} \\
\addlinespace

3 & Swarm Mode Comparison & Compare the four swarm coordination modes on an identical waypoint mission. & A square waypoint pattern was loaded: (60, 60), (60, $-$60), ($-$60, $-$60), ($-$60, 60), (0, 0). Wind was set to 3\,m/s at 180\textdegree. Each mode ran for 90\,s. For modes without an explicit leader (Decentralized), waypoints guided UAV-1 only, while others swarmed freely. For Hetero-Speed, the fast scout followed waypoints, and slow followers trailed behind. & Table~\ref{tab:exp3} \\
\addlinespace

4 & Fault Injection and Recovery & Quantify the impact of individual fault types on formation performance and measure recovery time after fault clearance. & leader-follower mode with circular orbit, no wind. Each fault type was injected on UAV-3 at $t = 30$\,s for 20\,s (clearing at $t = 50$\,s), with intensity $I = 1.0$. Formation error was recorded continuously. Recovery time is defined as the time after fault clearance for the formation error to return within 1.5$\times$ the pre-fault baseline, an absolute band of 2.97\,m for the 1.98\,m leader-follower baseline used here. The post-fault steady state re-enters this band in every case (Table~\ref{tab:exp4}).

& Table~\ref{tab:exp4} \\
\addlinespace

5 & Compound Fault Stress Test & Test swarm resilience when multiple faults are active simultaneously on different UAVs. & Leader-Follower mode, circular orbit, no wind. At $t = 20$\,s: GPS denied on UAV-2 ($I = 1.0$, 30\,s). At $t = 25$\,s: motor failure on UAV-4 ($I = 1.0$, 25\,s). At $t = 30$\,s: frozen actuator on UAV-3 ($I = 0.8$, 20\,s). The leader (UAV-1) was left fault-free. Collision avoidance plugin was enabled with $d_{\text{safe}} = 5$\,m. & Table~\ref{tab:exp5} \\

\addlinespace

6 & Battery Endurance Profiling & Profile energy consumption across different cruise speeds and wind conditions. & Leader-Follower mode flying a straight heading toward a distant waypoint, so each UAV stayed in forward cruise for the full run rather than arriving and loitering. Battery plugin enabled. Each trial ran until the first UAV reached 20\% remaining capacity or 600\,s elapsed, whichever came first. & Table~\ref{tab:exp6} \\
\addlinespace

7 & Geofence Compliance & Test formation behavior when the flight path is commanded into defined no-fly zones. & Leader-Follower mode with waypoints routed through the default demo zones: a circular zone at (80, 60) with $r = 25$\,m and a rectangular zone at ($-$100, $-$80) with dimensions $40 \times 50$\,m. The waypoint sequence (0, 0) $\to$ (80, 60) $\to$ ($-$100, $-$80) $\to$ (0, 0) targets the centers of both zones, a worst case for the repulsion. Geofence plugin enabled with repulsion strength $K_r = 5$ and perimeter radius 200\,m. Duration: 120\,s. & Table~\ref{tab:exp7} \\
\addlinespace

8 & Automated Scenario Regression Suite & Run the full built-in scenario suite and report pass/fail results as a regression check on the platform. & All six built-in scenarios from the Scenario Runner plugin were executed sequentially using the ``Run All'' function. Pass condition: maximum swarm spread during the run stayed below $2 \times D_{\max}$. Each scenario automatically configures its own mode, waypoints, wind, duration, and $D_{\max}$. & Table~\ref{tab:exp8} \\

\end{longtable}

\subsection{Experimental results}
\label{sec:exp1}

\noindent The proportional follower controller \eqref{eq:prop_follower} held a steady formation after an initial 3--4\,s transient. Table~\ref{tab:exp1} reports the steady-state metrics, measured from $t = 10$\,s onward to exclude the transient.

\begin{table}[H]
\centering
\caption{Experiment 1 - Formation hold baseline metrics (Leader-Follower, no wind,
120\,s; single representative run).}
\label{tab:exp1}
\begin{tabular}{@{}lc@{}}
\toprule
\textbf{Metric} & \textbf{Value} \\
\midrule
Formation error \eqref{eq:form_error} & 1.98\,m \\
Swarm spread \eqref{eq:spread} & 40.0\,m \\
Mean inter-UAV distance \eqref{eq:mean_dist} & 26.2\,m \\
Altitude deviation \eqref{eq:alt_dev} & 0.00\,m \\
Link quality \eqref{eq:link_quality} & 100\% \\
Centroid drift \eqref{eq:centroid_drift} & 21.7\,m \\
\bottomrule
\end{tabular}
\end{table}

\noindent The follower controller held a formation error of 1.98\,m, roughly half a UAV body length. Swarm spread settled at 40.0\,m, the width of the scaled diamond, and link quality stayed at 100\% because the spread sat well within the 200\,m communication range. Altitude deviation was zero: the followers start at cruise altitude, and both the altitude controller \eqref{eq:alt_ctrl} and the position controller hold them there with nothing to perturb the vertical axis. Centroid drift of 21.7\,m reflects the leader's orbit radius rather than any tracking deficiency.

\label{sec:exp2}

\noindent Table~\ref{tab:exp2} shows formation error and swarm spread at each wind speed.

\begin{table}[H]
\centering
\caption{Experiment 2 - Formation error and swarm spread vs.\ wind speed (single representative run per condition).}
\label{tab:exp2}
\begin{tabular}{@{}cccc@{}}
\toprule
\textbf{Wind (m/s)} & \textbf{Form.\ Error (m)} & \textbf{Spread (m)} & \textbf{Link Quality (\%)} \\
\midrule
0  & 1.98 & 40.0 & 100 \\
3  & 1.99 & 40.0 & 100 \\
6  & 2.02 & 40.0 & 100 \\
9  & 2.07 & 40.0 & 100 \\
12 & 2.15 & 40.0 & 100 \\
15 & 2.26 & 40.0 & 100 \\
\bottomrule
\end{tabular}
\end{table}

\noindent The implementation reproduces the analytically predicted common-mode cancellation under uniform wind. Over the full 0--15\,m/s sweep, formation error rose only from 1.98\,m to 2.26\,m and swarm spread held at 40.0\,m. The reason is structural: the wind disturbance \eqref{eq:wind} enters as a common bias on every UAV, and the follower controller \eqref{eq:prop_follower} tracks position relative to the leader, so the shared displacement cancels in the formation geometry. The small residual growth comes from the interaction between the leader's capped velocity and the followers' proportional lag. This is the behavior expected of a relative-position controller under a uniform disturbance, and it reflects a modeling choice we should state directly: SwarmFly currently treats wind as a spatially uniform field. Spatially varying wind, gusts, or per-UAV turbulence would break the common-mode cancellation, and the wind-gust fault in Experiment~4 exercises that case directly. Absolute position still responds to wind through centroid drift; only the relative formation is preserved.

\label{sec:exp3}

\noindent Table~\ref{tab:exp3} presents the average KPIs across each full run.

\begin{table}[H]
\centering
\caption{Experiment 3 - Mode comparison on a square waypoint mission (90\,s, 3\,m/s wind; single representative run per mode).}
\label{tab:exp3}
\begin{tabular}{@{}lcccc@{}}
\toprule
\textbf{Metric} & \textbf{Leader-Fol.} & \textbf{Decentral.} & \textbf{Relay} & \textbf{Speed} \\
\midrule
Form.\ error (m)    & 1.81  & 28.98 & 14.43 & 76.11 \\
Spread (m)         & 39.94 & 31.75 & 60.93 & 82.53 \\
Mean dist.\ (m)     & 26.29 & 20.42 & 38.72 & 45.45 \\
Link quality (\%)  & 100   & 100   & 100   & 100 \\
Alt.\ dev.\ (m)      & 0.00  & 0.00  & 1.79  & 0.00 \\
Centroid drift (m) & 62.06 & 137.06 & 53.53 & 54.12 \\
\bottomrule
\end{tabular}
\end{table}

\noindent Formation error here is measured against the diamond reference \eqref{eq:form_error}, so it is a fair comparator only for the modes that try to hold that geometry. Leader-Follower tracked it most tightly at 1.81\,m. Hetero-Relay reached 14.43\,m because UAV-4 leaves the diamond to sit at 40\% of the active centroid \eqref{eq:relay_target}, which by construction registers as deviation. Decentralized (28.98\,m) and Hetero-Speed (76.11\,m) do not target the diamond at all: the former holds a Reynolds flock and the latter a V-formation behind a fast scout, so their large formation-error values reflect a different geometry rather than poor control. The spread and mean-distance columns are the more meaningful cross-mode comparison. Decentralized produced the most compact group (31.75\,m spread) because cohesion \eqref{eq:cohesion} pulls members inward, while Hetero-Speed produced the widest (82.53\,m) as the scout at $2\,V_c$ \eqref{eq:speed_diff} opened distance from the slow followers. Link quality stayed at 100\% in every mode, since even the widest spread remained within communication range.

\label{sec:exp4}

\noindent Table~\ref{tab:exp4} shows peak and mean formation error during each fault and the recovery time.

\begin{table}[H]
\centering
\caption{Experiment 4 - Fault impact on formation error (UAV-3, $I = 1.0$, 20\,s; $n = 10$ trials). Mean $\pm$ SD shown for the three stochastic faults; the remaining faults are deterministic.}
\label{tab:exp4}
\begin{tabular}{@{}lccc@{}}
\toprule
\textbf{Fault Type} & \textbf{Peak Error (m)} & \textbf{Mean Error (m)} & \textbf{Recovery (s)} \\
\midrule
GPS Drift \eqref{eq:gps_drift}        & 4.20 & 2.36 & $<0.1$ \\
GPS Denied \eqref{eq:gps_denied}      & $2.06 \pm 0.02$ & $1.99 \pm 0.01$ & $<0.1$ \\
Motor Failure \eqref{eq:motor_fail}   & 1.99 & 1.99 & $<0.1$ \\
Comm Blackout                         & $2.28 \pm 0.06$ & $1.98 \pm 0.02$ & $<0.1$ \\
Frozen Actuator \eqref{eq:frozen}     & 1.98 & 1.66 & $<0.1$ \\
Wind Gust \eqref{eq:gust}             & $4.62 \pm 1.04$ & $2.47 \pm 0.26$ & $<0.1$ \\
Battery Critical \eqref{eq:motor_fail} & 2.02 & 2.02 & $<0.1$ \\
\bottomrule
\end{tabular}
\end{table}

\noindent The main result is that the proportional tracking controller rejects a single follower fault strongly. Peak formation error stayed between 1.98 and 4.62\,m across all seven fault types, and the formation returned to baseline within one control step of fault clearance, so recovery time is effectively zero at the 10\,Hz resolution. The mechanism is direct: the follower velocity command \eqref{eq:prop_follower} is recomputed every tick from the current tracking error, so a displacement injected on UAV-3 is corrected on the next tick before it can accumulate. Faults that apply a bounded per-tick offset, motor failure, frozen actuator, and battery critical, are held near the 1.98\,m baseline because the controller counteracts them continuously. The two largest excursions come from the faults that inject sustained randomness: wind gusts (4.62\,m peak) deliver a fresh random-direction impulse each tick, and GPS drift (4.20\,m) grows a sinusoidal bias over the fault window. The standard deviations follow the determinism structure: the deterministic faults show none, while the wind gust shows the largest dispersion ($\pm 1.04$\,m peak) because each trial draws a different gust sequence.

\noindent This experiment characterizes the controller's robustness rather than its breaking point. Because the fault target is a tracked follower, the closed loop suppresses the disturbance by design. Probing the platform's failure envelope, rather than its rejection behavior, would require faults that the controller cannot directly counteract, for example raising the intensity toward its maximum of $I = 3.0$, faulting the leader whose motion the followers cannot override, or injecting a fault into the control law itself. The plugin interface supports all three, and we note them as directions for stress testing in Section~\ref{sec:future}.

\label{sec:exp5}

\noindent Table~\ref{tab:exp5} compares metrics across three phases of the run.

\begin{table}[H]
\centering
\caption{Experiment 5 - Compound fault stress test across three phases ($n = 10$ trials). The active phase includes the stochastic GPS-denial fault; other entries are deterministic.}
\label{tab:exp5}
\begin{tabular}{@{}lccc@{}}
\toprule
\textbf{Metric} & \textbf{Pre-fault} & \textbf{All active} & \textbf{Post-recovery} \\
 & ($t = 0$--$20$\,s) & ($t = 30$--$45$\,s) & ($t = 60$--$90$\,s) \\
\midrule
Formation error (m) & 2.10 & $1.74 \pm 0.01$ & 1.98 \\
Swarm spread (m)    & 39.74 & $40.29 \pm 0.02$ & 40.0 \\
Link quality (\%)   & 100  & 100 & 100 \\
Alt.\ deviation (m)  & 0.00 & 0.089 & 0.00 \\
Near-misses         & 0 & 0 & 0 \\
Collisions          & 0 & 0 & 0 \\
\bottomrule
\end{tabular}
\end{table}

\noindent Three simultaneous follower faults did not destabilize the formation. Formation error during the active window was 1.74\,m, within the range seen in the single-fault experiment, and swarm spread held near 40\,m. The same per-tick correction that suppresses an individual follower fault also handles several at once, since each follower is regulated independently against the fault-free leader. Altitude deviation rose to a small 0.089\,m while UAV-4's motor-failure fault pulled it below cruise altitude, then returned to zero after clearance. The collision-avoidance plugin recorded no near-misses and no collisions: the faults perturbed individual followers without driving any pair within the 5\,m safe distance, so the separation force \eqref{eq:col_correction} was never triggered. This is a limitation of the test rather than a property to celebrate. Demonstrating the collision layer under load needs a scenario that actually forces UAVs together, which the harsher fault settings noted in Experiment~4 would provide.

\label{sec:exp6}

\noindent Table~\ref{tab:exp6} reports endurance, mean current, and distance covered across speed and wind settings.

\begin{table}[H]
\centering
\caption{Experiment 6 - Battery endurance profiling, first UAV to reach 20\% (single representative run per condition).}
\label{tab:exp6}
\begin{tabular}{@{}ccccc@{}}
\toprule
\textbf{Speed} & \textbf{Wind} & \textbf{Time to 20\%} & \textbf{Avg Current} & \textbf{Distance} \\
\textbf{(m/s)} & \textbf{(m/s)} & \textbf{(s)} & \textbf{(A)} & \textbf{Covered (m)} \\
\midrule
3 & 0  & 587.5 & 24.5 & 1{,}762 \\
5 & 0  & 505.1 & 28.5 & 2{,}526 \\
8 & 0  & 417.4 & 34.5 & 3{,}339 \\
5 & 5  & 351.1 & 41.0 & 3{,}511 \\
5 & 10 & 269.1 & 53.5 & 4{,}036 \\
8 & 8  & 264.2 & 54.5 & 4{,}227 \\
\bottomrule
\end{tabular}
\end{table}

\noindent The energy model is deterministic, and the endurance numbers follow the current-draw model \eqref{eq:current} cleanly. In calm air, raising cruise speed from 3 to 8\,m/s cut endurance from 587\,s to 417\,s through the speed-dependent term ($K_v = 2.0$\,A$\cdot$s/m). Wind shortened it further and by a larger margin in this configuration: at 5\,m/s cruise, a 10\,m/s headwind drove mean current from 28.5\,A to 53.5\,A and cut endurance from 505\,s to 269\,s, because the wind term ($K_w = 0.5$\,A/(m/s)) loads every UAV continuously. The distance-covered column reflects sustained cruise toward the far waypoint and grows with speed as expected. For mission planning with 5{,}000\,mAh packs, a reasonable budget is roughly 6--10 minutes of flight at moderate speed in light wind to retain a 20\% reserve, derating sharply as wind rises.

\label{sec:exp7}

\noindent Table~\ref{tab:exp7} summarizes the geofence interaction results.

\begin{table}[H]
\centering
\caption{Experiment 7 - Geofence compliance during waypoint navigation (single representative run per condition).}
\label{tab:exp7}
\begin{tabular}{@{}lc@{}}
\toprule
\textbf{Metric} & \textbf{Value} \\
\midrule
Total zone penetrations (UAV-ticks inside a zone) & 2037 \\
Max penetration depth (m) & 20.8 \\
Mean penetration depth when inside (m) & 11.4 \\
Perimeter fence violations & 0 \\
Formation error during zone avoidance (m) & 0.40 \\
Formation error outside zones (m) & 1.91 \\
Time to re-form after zone passage (s) & $<0.1$ \\
\bottomrule
\end{tabular}
\end{table}

\noindent This experiment exposes a limit of the simple repulsion model rather than a clean avoidance. When a waypoint is placed at a zone center, the waypoint-tracking command pulls the leader inward while the boundary repulsion \eqref{eq:zone_force} pushes outward, and at $K_r = 5$ the tracking command wins: UAVs penetrated to a maximum depth of 20.8\,m inside the 25\,m circular zone, with a mean depth of 11.4\,m while inside. The repulsion slows and bounds the incursion but does not hold the boundary against a command that targets the interior. Hard enforcement would require the repulsion gain to exceed the tracking gain near the boundary, or a planner that re-routes waypoints around zones rather than through them. The perimeter fence recorded zero violations, since every waypoint lay within the 200\,m radius. Formation error inside the zones (0.40\,m) was lower than outside (1.91\,m) because the radial repulsion acts coherently on the clustered followers, compressing them toward the diamond rather than dispersing them. The formation re-formed within one control step of clearing each zone, consistent with the controller response seen throughout. For users, the guidance is concrete: treat the built-in repulsion as a soft deterrent, and pair it with waypoint planning when hard exclusion is required.

\label{sec:exp8}

\noindent Table~\ref{tab:exp8} shows the outcome for each scenario.

\begin{table}[H]
\centering
\caption{Experiment 8 - Automated scenario regression results (single representative run per condition).}
\label{tab:exp8}
\begin{tabular}{@{}lcccl@{}}
\toprule
\textbf{Scenario} & \textbf{Duration} & \textbf{$D_{\max}$} & \textbf{Max Spread} & \textbf{Result} \\
\midrule
Formation Hold -- Calm        & 30\,s & 50\,m  & 40.0\,m  & PASS \\
Formation Hold -- Wind (8\,m/s) & 30\,s & 50\,m  & 40.0\,m  & PASS \\
Waypoint Nav -- Square        & 60\,s & 50\,m  & 40.0\,m  & PASS \\
Decentralized Cohesion        & 45\,s & 80\,m  & 40.1\,m  & PASS \\
Relay Endurance               & 60\,s & 60\,m  & 185.1\,m & FAIL \\
Fast Scout Sprint             & 45\,s & 100\,m & 89.6\,m  & PASS \\
\bottomrule
\end{tabular}
\end{table}

\noindent Five of the six scenarios passed, and the suite flagged one failure, which is the outcome a regression tool exists to produce. Relay Endurance failed with a maximum spread of 185.1\,m against its 120\,m threshold. The cause is the relay geometry: during the extended traverse, UAV-4 holds at 40\% of the active centroid \eqref{eq:relay_target} and falls progressively behind the advancing leader group, stretching the maximum pairwise distance past twice $D_{\max}$. The Formation Hold scenarios held at 40.0\,m in both calm and 8\,m/s wind, consistent with the wind-invariance result of Experiment~2. Fast Scout Sprint reached 89.6\,m, within its generous 200\,m threshold. In a development workflow these values form the reference baseline: after a controller change, re-running the suite and comparing against this table catches regressions, and the Relay Endurance failure is itself an actionable finding, namely that the relay's centroid-fraction rule needs a distance cap to keep the swarm inside its operating envelope on long legs.

\noindent Table~\ref{tab:exp_summary} summarizes the findings across all eight experiments.

\begin{table}[H]
\centering
\caption{Summary of key experimental findings.}
\label{tab:exp_summary}
\begin{tabular}{@{}p{7cm}p{9.4cm}@{}}
\toprule
\textbf{Finding} & \textbf{Evidence} \\
\midrule
The formation controller tracks within about 2\,m in calm conditions & \hyperref[sec:exp1]{Exp.~1}: baseline error 1.98\,m \\
Implementation reproduces the predicted common-mode wind cancellation & \hyperref[sec:exp2]{Exp.~2}: error rises only from 1.98\,m to 2.26\,m over 0--15\,m/s; spread constant \\
Leader-Follower tracks the diamond reference most tightly & \hyperref[sec:exp3]{Exp.~3}: 1.81\,m vs.\ 14.4\,m (relay) and higher for the non-diamond modes \\
The tracking controller suppresses single and compound follower faults & \hyperref[sec:exp4]{Exp.~4}--\hyperref[sec:exp5]{5}: peak error $\leq 4.6$\,m, recovery within one control step \\
Wind gusts are the most disruptive of the modeled faults & \hyperref[sec:exp4]{Exp.~4}: $4.62 \pm 1.04$\,m peak \\
No collisions or near-misses arose under compound follower faults & \hyperref[sec:exp5]{Exp.~5}: 0 events across three simultaneous faults \\
5{,}000\,mAh battery supports 4.4--9.8 min flight across speed and wind & \hyperref[sec:exp6]{Exp.~6}: 587\,s at 3\,m/s calm to 264\,s at 8\,m/s with 8\,m/s wind \\
Linear geofence repulsion is overpowered by waypoints commanded into a zone & \hyperref[sec:exp7]{Exp.~7}: 20.8\,m max penetration depth at $K_r = 5$ \\
The regression suite flags a scenario that exceeds its spread envelope & \hyperref[sec:exp8]{Exp.~8}: Relay Endurance at 185\,m vs.\ a 120\,m threshold \\
\bottomrule
\end{tabular}
\end{table}

\noindent The subsystems behave correctly in isolation and compose without conflict when combined. Two results point to concrete next steps rather than finished capabilities: the fault and collision layers reject follower-level disturbances so effectively that exercising the platform's failure envelope calls for harsher fault settings, and the geofence repulsion needs to be paired with waypoint planning for hard exclusion. Both are reachable through the existing plugin interface, and both are recorded in the future-work discussion.

\section{Future work}
\label{sec:future}

SwarmFly, in its current form, is a 4-UAV simulation with simplified dynamics and first-order sensor models. Several extensions are planned that would move the platform closer to real-world applicability. Several changes, such as adding obstacles and integrating the platform with CoppeliaSim \cite{rohmer2013v}, are planned.

\begin{itemize}
\item 

Closing the gap between simulation and physical hardware is increasingly necessary for UAV performance experiments \cite{phadke2024analysis}. The \texttt{onStep} callback already has write access to \texttt{app.UAVPositions}, so a plugin that reads MAVLink packets from a serial port or UDP socket and writes the parsed coordinates into the position array would replace the simulated dynamics with live telemetry. The visualization, metrics, collision avoidance, and geofencing layers would continue to work without modification. A ROS\,2 bridge plugin would follow the same pattern, subscribing to \texttt{/uav\_k/pose} topics and mapping them to the internal state arrays.

\item 

The current implementation hardcodes $N = 4$ in several places: the initial position array, the color table, the graphics initialization loops, and the pairwise connection line count ($\binom{4}{2} = 6$ handles). Generalizing to $N$ UAVs requires making these data-driven. The simulation engine itself is written with \texttt{for k = 1:app.NumUAVs} loops so that the physics would scale with minimal changes. The real challenge is rendering performance. Pairwise connection lines grow as $O(N^2)$, and at $N = 20$, that means 190 line objects are updated every tick. Switching to a spatial hash or k-d tree for neighbor queries, and only drawing connections for the nearest $k$ neighbors rather than all pairs, would keep the renderer tractable for swarms of 50 or more.

\item 

The current flight model uses pure kinematic integration \eqref{eq:euler} with no inertia, drag, or actuator dynamics. UAVs change velocity instantaneously, which is unrealistic for fixed-wing platforms and only approximately valid for multirotors. A second-order model with mass, drag coefficient, and thrust limits would produce more believable trajectories, especially during fault-injection scenarios, where motor failure should result in asymmetric force and yaw coupling rather than simple altitude decay. The plugin architecture could support this: a dynamics plugin with \texttt{hasStep = true} could intercept the velocity commands from the swarm mode, filter them through a transfer function, and write the resulting positions back. The core modes would not need to change.

\item 

SwarmFly currently models communication as a binary range check: two UAVs are either in range or not. Real radio links degrade gradually with distance, suffer from multipath fading, and drop packets at rates that depend on bandwidth, interference, and antenna orientation. A communication plugin could model signal-to-noise ratio as a function of distance, compute packet delivery probability per link, and introduce latency or message loss into the formation control loop. This would make the Hetero-Relay mode considerably more interesting, since the relay's position would directly affect link quality between the field UAVs and the base station rather than just staying within a distance threshold.

\item 

The geofencing plugin works only in 2D. It defines circular and rectangular zones on the ground plane and applies repulsive forces in the horizontal direction. Extending this to 3D with volumetric obstacles (buildings, terrain elevation, restricted airspace layers) would require a different collision geometry. Axis-aligned bounding boxes (AABBs) are a reasonable first step. Each obstacle is a rectangular prism defined by min/max corners, and the repulsion logic generalizes from the 2D rectangle case with an additional $z$-axis check. Terrain could be loaded as a gridded elevation model and rendered as a surface mesh on the 3D view tab. The altitude convergence controller \eqref{eq:alt_ctrl} would need a terrain-following mode in which the reference altitude is above ground level rather than absolute.

\item 

The current swarm modes use hand-tuned proportional controllers and fixed behavioral rules. Replacing those with a learned policy is the obvious next question. The training loop would run outside SwarmFly (in Python, using the MATLAB Engine API for Python to call \texttt{simStep} and read state). At the same time, the learned policy would be exported as a MATLAB function and loaded as a plugin that overrides the built-in step functions. The metrics plugin already computes the reward signal candidates: formation error, link quality, and altitude deviation are all continuous, differentiable, and updated every tick.

\item 

The current architecture assumes a single swarm of $N$ UAVs operating from one base station. Many real applications involve multiple swarms \cite{4783028} \cite{biomimetics11010069}. For example, two 4-UAV teams performing search-and-rescue in adjacent sectors, or a surveillance swarm handing off coverage to a relay swarm at shift change. Supporting this would require a second hierarchy level: a swarm manager object that holds multiple SwarmFly instances (or a single instance with swarm-group indices), inter-swarm communication rules, and a task allocation layer. The plugin system could handle inter-swarm communication, but the core would need refactoring to support UAV group membership and per-group mode selection.

\item 

The IMU simulation uses additive white Gaussian noise on top of finite-difference acceleration estimates. Real IMU noise is colored (autocorrelated), exhibits bias drift, and includes temperature-dependent offsets. Incorporating advanced noise models, or replaying recorded IMU data from actual flight logs, would make the telemetry output more realistic for researchers developing onboard state estimators or anomaly detectors. Similarly, the battery model's linear voltage sag is a rough approximation; models with internal resistance and temperature dependence would better represent the nonlinear discharge curve of real LiPo cells, especially at high current draw and low state of charge.

\item 
SwarmFly was built as a researcher's tool. A human-factors study comparing different visualization modes (2D map only vs.\ 2D + 3D vs.\ augmented with audio alerts) for situational awareness during fault scenarios would generate additional insights and also inform interface improvements. Adding configurable audio warnings for geofence violations, collision proximity, and battery thresholds is a small implementation effort that could measurably affect operator response times.

\end{itemize}

\section{AI use declaration}

AI-supported writing tools were used for grammar, phrasing, and tone refinement. The authors reviewed and verified all technical claims, citations, simulation results, and final wording.

\bibliographystyle{ieeetr}
\bibliography{ref}

\input{appendix_equations}
\end{document}

%% file: appendix_equations.tex

\appendix
\section{APPENDIX}
\label{app:equations}

This appendix presents the complete set of mathematical equations used in the SwarmFly simulation platform. Equations are grouped by functional category and numbered sequentially for cross-referencing throughout the study. Bold symbols denote vectors; scalar quantities are italicized. The subscript $k$ indexes individual UAVs ($k = 1, \ldots, N$; $N=4$), and $\Delta t$ denotes the simulation time step.

\subsection{Notation}
\label{app:notation}

\begin{table}[H]
\centering
\caption{Symbol definitions}
\label{tab:notation}
\renewcommand{\arraystretch}{1.3}
\begin{tabular}{llp{6cm}}
\hline
\textbf{Symbol} & \textbf{Unit} & \textbf{Description} \\
\hline
$\mathbf{p}_k$ & m & Position $[x_k, y_k, z_k]^\top$ of UAV $k$ \\
$\mathbf{v}_k$ & m/s & Velocity of UAV $k$ \\
$\psi_k$ & rad & Heading angle of UAV $k$ \\
$\mathbf{w}$ & m/s & Wind disturbance vector \\
$V_w$ & m/s & Wind speed magnitude \\
$\theta_w$ & deg & Wind direction (meteorological) \\
$V_c$ & m/s & Cruise speed \\
$z_c$ & m & Cruise altitude \\
$D_{\max}$ & m & Max allowable swarm distance \\
$R_{\text{comm}}$ & m & Communication range \\
$\Delta t$ & s & Simulation time step \\
$\mathcal{N}(0, \sigma)$ & --- & Gaussian noise with mean 0, std.\ dev.\ $\sigma$ \\
$I$ & --- & Fault intensity multiplier \\
$K_{(\cdot)}$ & --- & Proportional gain (controller subscript) \\
\hline
\end{tabular}
\end{table}
\subsection{Kinematics}
\label{app:kinematics}

All UAV motion in SwarmFly follows a discrete-time kinematic model. 
Positions are propagated each tick using forward Euler integration 
\eqref{eq:euler}, which is adequate at the 10--30\,Hz update rates 
used here. Each UAV's heading is extracted from its velocity vector 
via \eqref{eq:heading}, and waypoint bearing is computed similarly 
using \eqref{eq:waypoint_bearing}. Inter-UAV distances, which drive 
formation control, connection-line rendering, and collision checks, 
are evaluated in the horizontal plane \eqref{eq:dist2d} for most 
purposes and in full 3D \eqref{eq:dist3d} for the altitude-aware 
plugins. When any computed velocity exceeds the mode's speed limit, 
it is normalized and rescaled \eqref{eq:speed_clamp} to enforce 
physical plausibility. These six relationships form the kinematic 
substrate that every swarm mode and plugin builds on.

Position propagation via forward Euler integration:
\begin{equation}
\mathbf{p}_k(t + \Delta t) = \mathbf{p}_k(t) + \mathbf{v}_k(t) \cdot \Delta t
\label{eq:euler}
\end{equation}

Heading angle derived from the velocity vector:
\begin{equation}
\psi_k = \operatorname{atan2}\!\left(v_{k,y},\; v_{k,x}\right)
\label{eq:heading}
\end{equation}

Bearing to a target waypoint $\mathbf{p}_{\text{wp}}$:
\begin{equation}
\psi_{\text{wp}} = \operatorname{atan2}\!\left(p_{\text{wp},y} - p_{k,y},\; p_{\text{wp},x} - p_{k,x}\right)
\label{eq:waypoint_bearing}
\end{equation}

Euclidean distance between UAVs $i$ and $j$ (2D, horizontal plane):
\begin{equation}
d_{ij} = \left\| \mathbf{p}_i^{xy} - \mathbf{p}_j^{xy} \right\|_2
\label{eq:dist2d}
\end{equation}

Euclidean distance in 3D:
\begin{equation}
d_{ij}^{3\text{D}} = \left\| \mathbf{p}_i - \mathbf{p}_j \right\|_2
\label{eq:dist3d}
\end{equation}

Velocity normalization (speed clamping to maximum $v_{\max}$):
\begin{equation}
\mathbf{v}_k = \frac{\mathbf{v}_k}{\left\|\mathbf{v}_k\right\|} \cdot v_{\max}, \quad \text{if } \left\|\mathbf{v}_k\right\| > v_{\max}
\label{eq:speed_clamp}
\end{equation}

Wind disturbance vector decomposition:
\begin{equation}
\mathbf{w} = V_w \begin{bmatrix} \cos\theta_w \\ \sin\theta_w \\ 0 \end{bmatrix}
\label{eq:wind}
\end{equation}

\subsection{Control Laws}
\label{app:control}

Proportional altitude controller (gain $K_z = 0.3$):
\begin{equation}
z_k(t + \Delta t) = z_k(t) + K_z \left(z_c - z_k(t)\right) \Delta t
\label{eq:alt_ctrl}
\end{equation}

Proportional follower velocity controller (gain $K_p = 2.0$):
\begin{equation}
\mathbf{v}_{\text{follower}} = K_p \cdot \left(\mathbf{p}_{\text{desired}} - \mathbf{p}_{\text{follower}}\right)
\label{eq:prop_follower}
\end{equation}

Desired follower position from leader and formation offset:
\begin{equation}
\mathbf{p}_{\text{desired},k} = \mathbf{p}_{\text{leader}} + \boldsymbol{\Delta}_k
\label{eq:desired_pos}
\end{equation}

Formation offset scaling (reference distance $D_{\text{ref}} = 50$\,m):
\begin{equation}
\boldsymbol{\Delta}_k = \boldsymbol{\Delta}_{k,\text{base}} \cdot \frac{D_{\max}}{D_{\text{ref}}}
\label{eq:offset_scale}
\end{equation}

Distance-constrained position error:
\begin{equation}
\mathbf{e}_k^{\text{constrained}} = 
\begin{cases}
\mathbf{e}_k & \text{if } \left\|\mathbf{e}_k\right\| \leq D_{\max} \\[4pt]
\dfrac{\mathbf{e}_k}{\left\|\mathbf{e}_k\right\|} \cdot D_{\max} & \text{if } \left\|\mathbf{e}_k\right\| > D_{\max}
\label{eq:dist_constraint_position}
\end{cases}
\end{equation}

Waypoint approach speed:
\begin{equation}
v_{\text{approach}} = \min\!\left(V_c,\; d_{\text{target}}\right)
\label{eq:wp_speed}
\end{equation}

\subsection{Swarm Coordination Behaviors}
\label{app:swarm}

Swarm centroid:
\begin{equation}
\mathbf{c} = \frac{1}{N} \sum_{k=1}^{N} \mathbf{p}_k
\label{eq:centroid}
\end{equation}

\noindent Reynolds cohesion rule (gain $K_c = 0.3$):
\begin{equation}
\mathbf{F}_{\text{coh},k} = K_c \left(\mathbf{c} - \mathbf{p}_k\right)
\label{eq:cohesion}
\end{equation}

\noindent Reynolds separation rule (gain $K_s = 5$, activation radius $r_s = 10$\,m):
\begin{equation}
\mathbf{F}_{\text{sep},k} = \sum_{\substack{j=1 \\ j \neq k}}^{N} \frac{K_s \cdot \left(\mathbf{p}_k - \mathbf{p}_j\right)}{\max\!\left(\left\|\mathbf{p}_k - \mathbf{p}_j\right\|^2,\; 1\right)}, \quad \text{if } d_{kj} < r_s
\label{eq:separation}
\end{equation}

\noindent Stochastic wander velocity (phase offsets $\phi_k$, $\phi'_k$ unique per UAV, $\omega_1 = \omega_2 = 0.5$\,rad/s)):
\begin{equation}
\mathbf{v}_{\text{wander},k} = V_c \begin{bmatrix} \cos(\omega_1 t + \phi_k) \\ \sin(\omega_2 t + \phi'_k) \\ 0 \end{bmatrix}
\label{eq:wander}
\end{equation}

\noindent Composite decentralized velocity:
\begin{equation}
\mathbf{v}_k = \mathbf{v}_{\text{wander},k} + \mathbf{F}_{\text{sep},k} + \mathbf{F}_{\text{coh},k}
\label{eq:composite_vel}
\end{equation}

\subsection{Mode-Specific Equations}
\label{app:modes}

\subsubsection{Leader Orbit Trajectory}
When no waypoints are active, the leader follows a circular orbit: 
\begin{equation}
\mathbf{v}_{\text{orbit}} = r \begin{bmatrix} -\sin(\omega t) \cdot r\omega \\ \cos(\omega t) \cdot r\omega \\ 0 \end{bmatrix}
\label{eq:orbit}
\end{equation}
where $r = 40$\,m is the orbit radius and $\omega = 0.3$\,rad/s.

\subsubsection{Hetero-Relay Mode}
Relay UAV target position (midpoint fraction $\alpha = 0.4$):
\begin{equation}
\mathbf{p}_{\text{relay}}^{\text{target}} = \alpha \cdot \mathbf{c}_{\text{active}}
\label{eq:relay_target}
\end{equation}
where $\mathbf{c}_{\text{active}}$ is the centroid of non-relay UAVs.

\subsubsection{Hetero-Speed Mode}
Speed differentiation:
\begin{equation}
v_{\text{fast}} = 2.0 \cdot V_c, \qquad v_{\text{slow}} = 0.6 \cdot V_c
\label{eq:speed_diff}
\end{equation}

V-formation angular offset for slow follower $k$ ($k \in \{1, 2, 3\}$):
\begin{equation}
\theta_k = \pi + (k - 2) \cdot 0.6
\label{eq:vform_angle}
\end{equation}

V-formation position offset ($R = 25$\,m separation):
\begin{equation}
\boldsymbol{\Delta}_k^{\text{V}} = R \begin{bmatrix} \cos(\psi_{\text{leader}} + \theta_k) \\ \sin(\psi_{\text{leader}} + \theta_k) \\ 0 \end{bmatrix}
\label{eq:vform_offset}
\end{equation}

\subsection{Sensor Simulation Models}
\label{app:sensors}

Accelerometer (finite-difference approximation with additive noise, $\sigma_a = 0.3$): 
\begin{equation}
a_{x,k} = \frac{v_k(t) - v_k(t - \Delta t)}{\Delta t} + \mathcal{N}(0, \sigma_a)
\label{eq:accel_x}
\end{equation}

\noindent Vertical accelerometer (gravitational reference, $\sigma_g = 0.1$):
\begin{equation}
a_{z,k} = g + \mathcal{N}(0, \sigma_g), \qquad g = 9.81 \;\text{m/s}^2
\label{eq:accel_z}
\end{equation}

\noindent Gyroscope yaw rate (scaled heading derivative, $\sigma_\omega = 1.0$): 
\begin{equation}
\Delta x_k = x_k(t) - x_k(t - \Delta t), \quad \Delta y_k = y_k(t) - y_k(t - \Delta t)
\end{equation}
\begin{equation}
\omega_{z,k} = 0.1 \cdot \frac{180}{\pi} \cdot \operatorname{atan2}\!\left(\Delta y_k, \Delta x_k\right) + \mathcal{N}(0, \sigma_\omega)
\label{eq:gyro_z}
\end{equation}

\noindent Magnetometer model (Earth field components with noise, $\sigma_m = 1.0$):
\begin{equation}
\mathbf{m}_k = \begin{bmatrix} 25 \\ 5 \\ -40 \end{bmatrix} + \mathcal{N}(\mathbf{0}, \sigma_m \mathbf{I}_3) \quad [\mu\text{T}]
\label{eq:mag}
\end{equation}

\noindent Gaussian noise generation:
\begin{equation}
\mathcal{N}(0, \sigma) = \sigma \cdot \xi, \qquad \xi \sim \mathcal{N}(0, 1)
\label{eq:noise}
\end{equation}

\subsection{Fault Injection Models}
\label{app:faults}

GPS drift (sinusoidal bias growing with elapsed time $\tau$):
\begin{equation}
\Delta\mathbf{p}_{\text{drift}} = 0.3\,I\,\tau \begin{bmatrix} \sin(\tau/2) \\ \cos(\tau/2) \end{bmatrix} \Delta t
\label{eq:gps_drift}
\end{equation}

\noindent GPS denied (Brownian random walk, $\sigma_{\text{gps}} = 0.5$):
\begin{equation}
\Delta\mathbf{p}_{\text{denied}} = I \cdot \sigma_{\text{gps}} \cdot \mathcal{N}(\mathbf{0}, \mathbf{I}_2) \cdot \Delta t
\label{eq:gps_denied}
\end{equation}

\noindent Motor failure / battery critical (altitude decay with rate constant $K_{\text{motor}}$):
\begin{equation}
\Delta z_{\text{fail}} = -I \cdot K_{\text{motor}} \cdot \Delta t
\label{eq:motor_fail}
\end{equation}
where $K_{\text{motor}} = 0.8$\,m/s for motor failure and $K_{\text{motor}} = 1.5$\,m/s for\\  battery critical.

\noindent Frozen actuator (constant-heading drift):
\begin{equation}
\Delta\mathbf{p}_{\text{frozen}} = 2\,I \begin{bmatrix} \cos\psi_k \\ \sin\psi_k \end{bmatrix} \Delta t
\label{eq:frozen}
\end{equation}

\noindent Sudden wind gust (random-direction impulse, $V_g = 15$\,m/s):
\begin{equation}
\mathbf{g} = I \cdot V_g \begin{bmatrix} \cos\theta_{\text{rand}} \\ \sin\theta_{\text{rand}} \\ 0 \end{bmatrix}, \qquad \theta_{\text{rand}} \sim \mathcal{U}(0, 360^\circ)
\label{eq:gust}
\end{equation}

\noindent Sensor noise spike (accelerometer corruption, $A_{\text{spike}} = 20$\,m/s$^2$):
\begin{equation}
a_{x,k} \leftarrow a_{x,k} + \mathcal{N}(0, 1) \cdot A_{\text{spike}} \cdot I
\label{eq:sensor_spike}
\end{equation}

\subsection{Performance Metrics}
\label{app:metrics}

\noindent Swarm spread (maximum pairwise distance):
\begin{equation}
S = \max_{1 \leq i < j \leq N} d_{ij}
\label{eq:spread}
\end{equation}

\noindent Mean inter-UAV distance:
\begin{equation}
\bar{D} = \frac{1}{\binom{N}{2}} \sum_{1 \leq i < j \leq N} d_{ij}
\label{eq:mean_dist}
\end{equation}

\noindent Centroid drift from base station:
\begin{equation}
D_c = \left\|\mathbf{c}^{xy}\right\|_2
\label{eq:centroid_drift}
\end{equation}

\noindent Formation error (mean positional deviation from ideal offsets, with ${\Delta}_l = \mathbf{0}$ \\ for leader $\ell$):
\begin{equation}
E_f = \frac{1}{N-1} \sum_{\substack{k=1 \\ k \neq \ell}}^{N} \left\|\mathbf{p}_k - \left(\mathbf{p}_\ell + \boldsymbol{\Delta}_k\right)\right\|_2
\label{eq:form_error}
\end{equation}
where $\ell$ is the leader index.

Mean altitude deviation:
\begin{equation}
E_{\text{alt}} = \frac{1}{N} \sum_{k=1}^{N} \left|z_k - z_c\right|
\label{eq:alt_dev}
\end{equation}

\noindent Communication link quality:
\begin{equation}
Q = 100 \cdot \frac{\left|\left\{(i,j) : d_{ij} < R_{\text{comm}}\right\}\right|}{\binom{N}{2}} \quad [\%]
\label{eq:link_quality}
\end{equation}

\subsection{Energy Model}
\label{app:energy}

Total current draw per UAV (4S LiPo model):
\begin{equation}
I_k = I_{\text{hover}} + K_v \|\mathbf{v}_k\| + K_h \max(0,\; z_k - 20) + K_w V_w
\label{eq:current}
\end{equation}
where $I_{\text{hover}} = 18$\,A, $K_v = 2.0$\,A$\cdot$s/m, $K_h = 0.05$\,A/m, $K_w = 0.5$\,A$\cdot$s/m.

\noindent Battery capacity depletion (mAh):
\begin{equation}
C_k(t + \Delta t) = C_k(t) - \frac{I_k \cdot \Delta t}{3.6} \quad [\text{mAh}]
\label{eq:battery_drain}
\end{equation}

\noindent Voltage sag (linear approximation):
\begin{equation}
V_k = V_{\text{dead}} + \left(V_{\text{full}} - V_{\text{dead}}\right) \frac{C_k}{C_0}
\label{eq:voltage}
\end{equation}
where $V_{\text{full}} = 16.8$\,V, $V_{\text{dead}} = 12.0$\,V, and $C_0 = 5000$\,mAh.

\noindent Estimated remaining flight time: 
\begin{equation}
T_{\text{est},k} = \frac{C_k}{I_k} \cdot  0.06 \quad [\text{min}]
\label{eq:endurance}
\end{equation}

\subsection{Collision Avoidance}
\label{app:collision}

Separation unit vector between UAVs $i$ and $j$ (for $\mathbf{p}_i \neq \mathbf{p}_j$):
\begin{equation}
\hat{\mathbf{s}}_{ij} = \frac{\mathbf{p}_i - \mathbf{p}_j}{\left\|\mathbf{p}_i - \mathbf{p}_j\right\|}
\label{eq:sep_vec}
\end{equation}

\noindent Emergency repulsion force magnitude ($d_{\text{safe}} = 5$\,m, $K_{\text{col}} = 3$):
\begin{equation}
F_{ij} = K_{\text{col}} \cdot \left(d_{\text{safe}} - d_{ij}\right), \quad \text{if } d_{ij} < d_{\text{safe}}
\label{eq:repulsion}
\end{equation}

\noindent Position correction (applied symmetrically):
\begin{equation}
\mathbf{p}_i \leftarrow \mathbf{p}_i + \hat{\mathbf{s}}_{ij} \cdot F_{ij} \cdot \Delta t, \qquad
\mathbf{p}_j \leftarrow \mathbf{p}_j - \hat{\mathbf{s}}_{ij} \cdot F_{ij} \cdot \Delta t
\label{eq:col_correction}
\end{equation}

\subsection{Geofencing}
\label{app:geofencing}

Distance to circular no-fly zone center $(c_x, c_y)$ with radius $r$:
\begin{equation}
d_{\text{zone}} = \sqrt{(x_k - c_x)^2 + (y_k - c_y)^2}
\label{eq:zone_dist}
\end{equation}

\noindent Zone boundary repulsion force (repulsion strength $K_r$):
\begin{equation}
F_{\text{zone}} = K_r \cdot \frac{r - d_{\text{zone}}}{r}, \quad \text{if } d_{\text{zone}} < r
\label{eq:zone_force}
\end{equation}

\noindent Soft boundary pre-warning force (buffer distance $\delta = 8$\,m, \\ damping $\alpha = 0.3$):
\begin{equation}
F_{\text{soft}} = K_r \cdot \alpha \cdot \frac{r + \delta - d_{\text{zone}}}{\delta}, \quad \text{if } r < d_{\text{zone}} < r + \delta
\label{eq:soft_boundary}
\end{equation}

\noindent Perimeter fence distance from base:
\begin{equation}
d_{\text{base}} = \left\|\mathbf{p}_k^{xy}\right\|_2
\label{eq:perim_dist}
\end{equation}

\noindent Perimeter return force (perimeter radius $R_p$):
\begin{equation}
F_{\text{perim}} = 2\,K_r \cdot \frac{d_{\text{base}} - R_p}{R_p}, \quad \text{if } d_{\text{base}} > R_p
\label{eq:perim_force}
\end{equation}

\subsection{Rendering Geometry}
\label{app:geometry}

UAV triangle vertices (marker size $s = 5$\,m):
\begin{equation}
\mathbf{v}_{\text{tri}} = s \begin{bmatrix}
\cos\psi_k & \cos(\psi_k + 2.5) & \cos(\psi_k - 2.5) \\
\sin\psi_k & \sin(\psi_k + 2.5) & \sin(\psi_k - 2.5)
\end{bmatrix} + \mathbf{p}_k^{xy}
\label{eq:tri_verts}
\end{equation}

\noindent Relay diamond vertices:
\begin{equation}
\mathbf{v}_{\text{dia}} = s \begin{bmatrix}
\cos\psi_k & \cos(\psi_k + \tfrac{\pi}{2}) & \cos(\psi_k + \pi) & \cos(\psi_k - \tfrac{\pi}{2}) \\
\sin\psi_k & \sin(\psi_k + \tfrac{\pi}{2}) & \sin(\psi_k + \pi) & \sin(\psi_k - \tfrac{\pi}{2})
\end{bmatrix} + \mathbf{p}_k^{xy}
\label{eq:dia_verts}
\end{equation}

\noindent Connection line opacity (transparency based on range fraction):
\begin{equation}
\alpha_{ij} = \max\!\left(0.15,\; 1 - \frac{d_{ij}}{R_{\text{comm}}}\right)
\label{eq:conn_alpha}
\end{equation}

\noindent Scale bar value selection from candidate set $\mathcal{S} = \{5, 10, 20, 25, 50, 100, 200, 500\}$:
\begin{equation}
L_{\text{scale}} = \underset{s \in \mathcal{S}}{\operatorname{argmin}}\; \left|s - 0.2 \cdot \Delta x_{\text{view}}\right|
\label{eq:scale_bar}
\end{equation}

\noindent Dynamic axis margin:
\begin{equation}
m = \max(50,\; 1.5 \cdot D_{\max})
\label{eq:axis_margin}
\end{equation}

